\documentclass[10pt,twocolumn,letterpaper]{article}

\usepackage[pagenumbers]{cvpr} 

\usepackage{graphicx}
\usepackage{amsmath}
\usepackage{amssymb}
\usepackage{booktabs}
\usepackage[accsupp]{axessibility}

\usepackage[pagebackref,breaklinks,colorlinks]{hyperref}

\usepackage{diagbox}
\usepackage{pifont}
\usepackage{threeparttable}

\newtheorem{theorem}{Theorem}
\newtheorem{corollary}{Corollary}[theorem]

\usepackage[capitalize]{cleveref}
\crefname{section}{Sec.}{Secs.}
\Crefname{section}{Section}{Sections}
\Crefname{table}{Table}{Tables}
\crefname{table}{Tab.}{Tabs.}

\newcommand*{\dif}{\mathop{}\!\mathrm{d}}

\def\cvprPaperID{3468} 
\def\confName{CVPR}
\def\confYear{2022}

\begin{document}

\title{Adversarial Texture for Fooling Person Detectors in the Physical World}

\author{Zhanhao Hu$^{1}$ \ \ Siyuan Huang$^{1}$ \ \ Xiaopei Zhu$^{2, 1}$ \ \ Fuchun Sun$^{1}$ \ \ Bo Zhang$^{1}$ \ \ Xiaolin Hu$^{1,3,4}$\thanks{Corresponding author.}\\
$^{1}$Department of Computer Science and Technology, Institute for Artificial Intelligence, \\ State Key Laboratory of Intelligent Technology and Systems, BNRist, Tsinghua University, Beijing, China \\
$^{2}$School of Integrated Circuits, Tsinghua University, Beijing, China \\
$^{3}$IDG/McGovern Institute for Brain Research, Tsinghua University, Beijing, China \\
$^{4}$Chinese Institute for Brain Research (CIBR), Beijing, China\\
\tt\small \{huzhanha17, zxp18\}@mails.tsinghua.edu.cn \\ \tt\small \{siyuanhuang, fcsun, dcszb, xlhu\}@mail.tsinghua.edu.cn}


\maketitle
\begin{abstract}
   Nowadays, cameras equipped with AI systems can capture and analyze images to detect people automatically. However, the AI system can make mistakes when receiving deliberately designed patterns in the real world, i.e., physical adversarial examples. Prior works have shown that it is possible to print adversarial patches on clothes to evade DNN-based person detectors. However, these adversarial examples could have catastrophic drops in the attack success rate when the viewing angle (i.e., the camera's angle towards the object) changes. To perform a multi-angle attack, we propose Adversarial Texture (AdvTexture). AdvTexture can cover clothes with arbitrary shapes so that people wearing such clothes can hide from person detectors from different viewing angles. We propose a generative method, named Toroidal-Cropping-based Expandable Generative Attack (TC-EGA), to craft AdvTexture with repetitive structures. We printed several pieces of cloth with AdvTexure and then made T-shirts, skirts, and dresses in the physical world. Experiments showed that these clothes could fool person detectors in the physical world.
\end{abstract}


\section{Introduction}
\label{sec:intro}
Recent works have shown that Deep Neural Networks (DNNs) are vulnerable to the adversarial examples crafted by adding subtle noise to the original images in the digital world ~\cite{goodfellow2014explaining,szegedy2013intriguing,kurakin2016adversarial,papernot2016the,nguyen2015deep,moosavi2016deepfool,carlini2017towards,dong2018boosting}, and that the DNNs can be attacked by manufactured objects in the physical world~\cite{sharif2016accessorize,athalye2018synthesizing,brown2017adversarial,evtimov2017robust}. These manufactured objects are called \emph{physical adversarial examples}. Recently, some methods based on patch attacks ~\cite{sharif2016accessorize} have been proposed to evade person detectors \cite{thys2019fooling,xu2020adversarial,huang2020universal,wu2020making,zhu_aaai,hu2021naturalistic}. Specifically, Thys et al. ~\cite{thys2019fooling} proposed to attach a patch to a cardboard. By holding the cardboard in front of the camera, the person cannot be detected by the person detectors. Xu et al. ~\cite{xu2020adversarial} proposed an adversarial T-shirt printed with adversarial patches. The person wearing the T-shirt can also evade person detectors. These works impose considerable threats to the widely deployed deep learning-based security systems. It urges researchers to re-evaluate the safety and reliability of these systems.




\begin{figure}[t]
  \centering
  \begin{subfigure}{0.132\textwidth}
  \includegraphics[width=\textwidth]{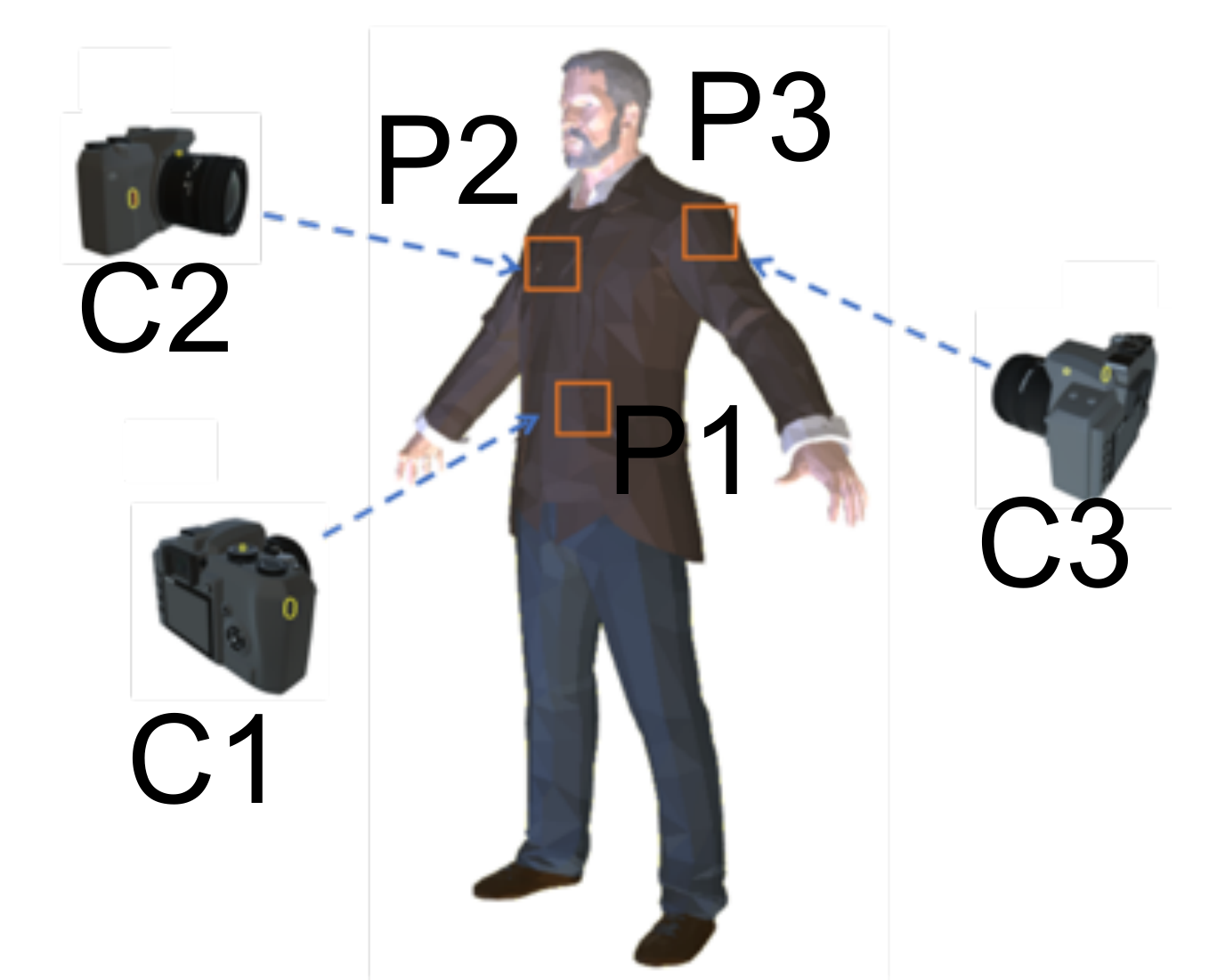}
  \caption{}
  \label{fig:3dperson}
  \end{subfigure}
  \begin{subfigure}{0.1\textwidth}
  \includegraphics[width=\textwidth]{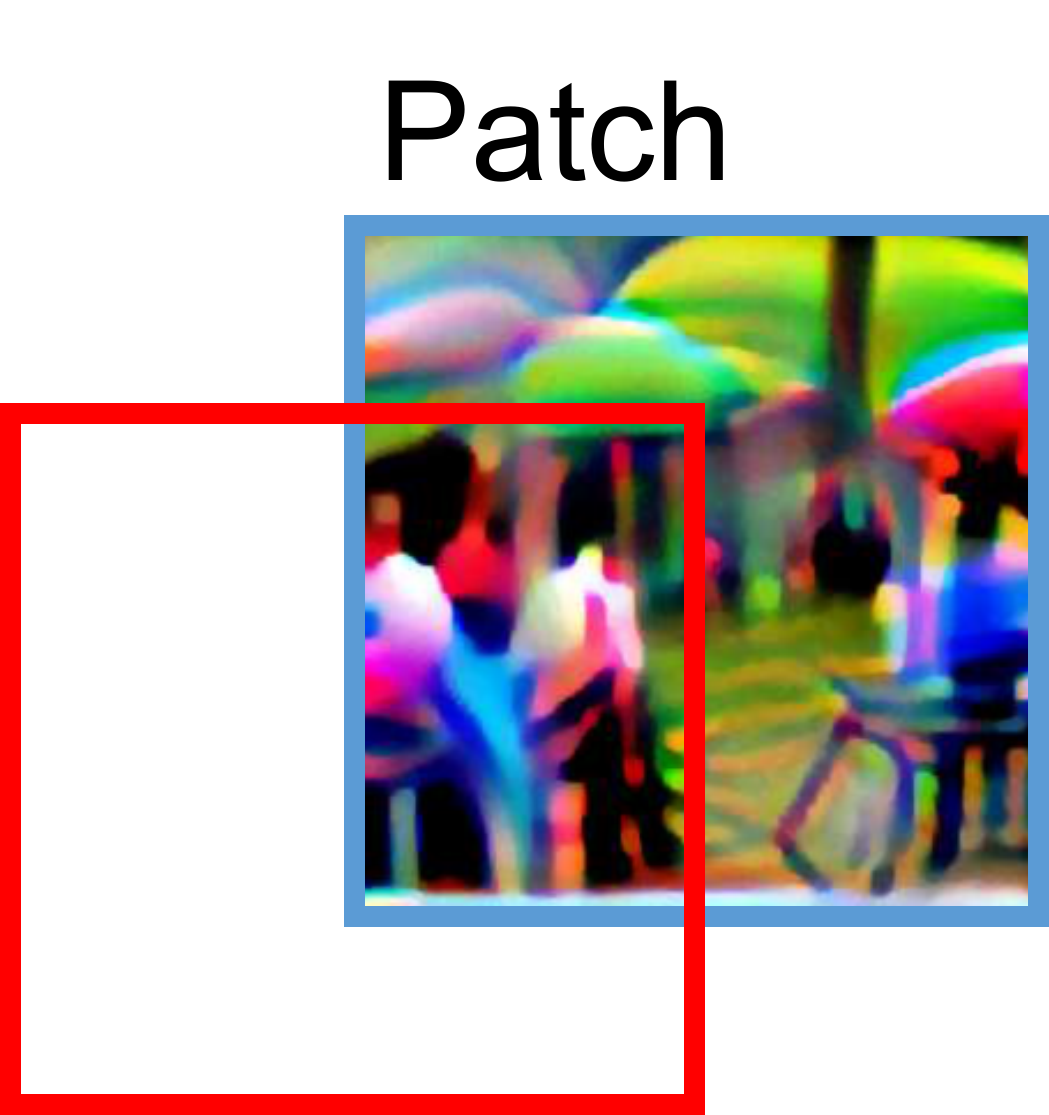}
  \caption{}
  \label{fig:diff1}
  \end{subfigure}
  \begin{subfigure}{0.1\textwidth}
  \includegraphics[width=\textwidth]{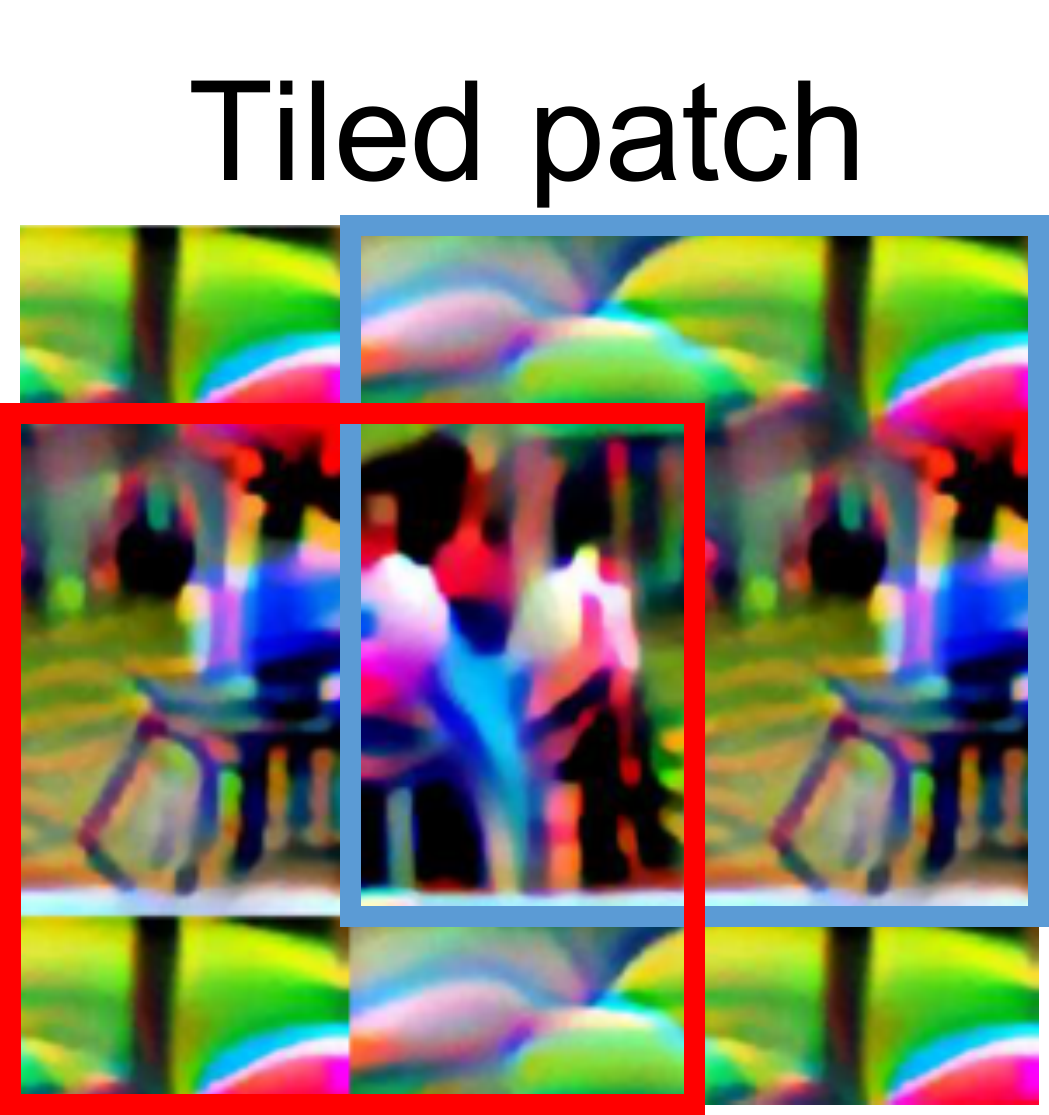}
  \caption{}
  \label{fig:diff2}
  \end{subfigure}
  \begin{subfigure}{0.1\textwidth}
  \includegraphics[width=\textwidth]{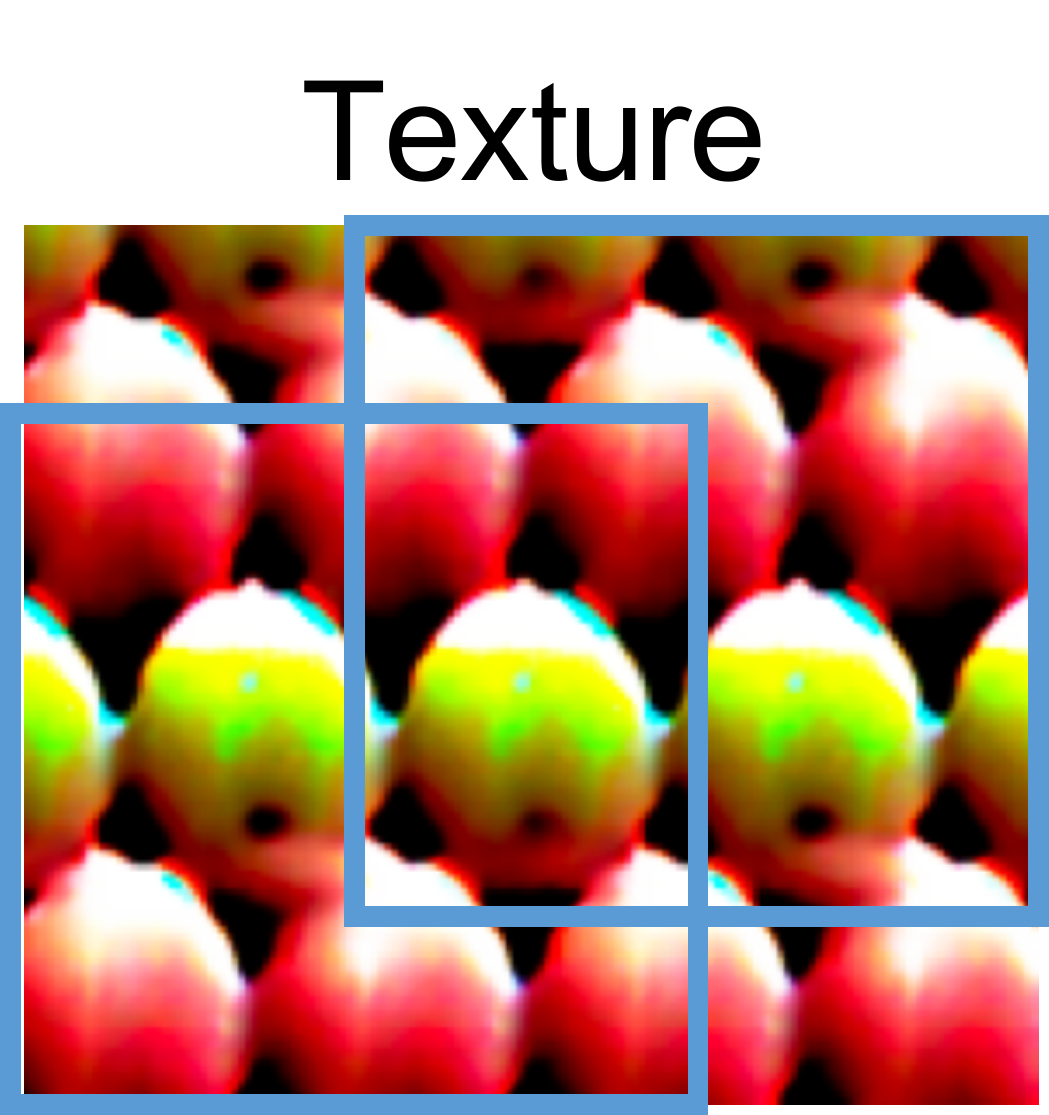}
  \caption{}
  \label{fig:diff3}
  \end{subfigure}
  \caption{Illustration of the attacks at different viewing angles. (a) The camera captures different parts (P1, P2, P3) of the clothes when set to different viewing angles (C1, C2, C3). (b-d) The boxes are the possible areas that the camera may capture. The blue ones indicate the most effective areas for attack, while the red ones are less effective.}
\end{figure}

However, the person detector attack methods mentioned above are effective only when the adversarial patches face the camera. Apparently, a single adversarial patch on a piece of clothing is hard to attack detectors at multiple viewing angles, as the camera may only capture a segment of the heavily deformed patch (\cref{fig:3dperson} and \cref{fig:diff1}). We call this the \emph{segment-missing} problem. A naive extension is to cover the clothing with multiple patches (e.g., tiling the patches tightly on the clothing; see \cref{fig:diff2}). However, it cannot totally solve the segment-missing problem, because the camera will capture several segments belonging to different patch units, making the attack inefficient. Another straightforward solution is to build a 3D model of a human body and a specific piece of clothing to render in different viewing angles as previous work ~\cite{athalye2018synthesizing} did. However, the clothes are non-rigid, and current 3D rendering techniques have difficulties in modeling the natural deformation of clothes in the real world. For example, Wang et al. \cite{wang2020can} rendered 3D logos on flat areas (front and back) of 3D human meshes, but the Attack Success Rate (ASR) decreased when applying to unseen meshes.



To solve the problem, we propose the idea of using Adversarial Texture (AdvTexture). Unlike the patch-based attacks, AdvTexture can be generated in arbitrary size, thus can cover any cloth in any size. We require that any local part of the texture has adversarial effectiveness (\cref{fig:diff3}). Then, when the clothes are covered with AdvTexture, every local area caught by the camera can attack the detectors, which solves the segment-missing problem.


Towards this goal, we propose a two-stage generative method, Toroidal-Cropping-based Expandable Generative Attack (TC-EGA), to craft AdvTexture. In the first stage, we train a fully convolutional network (FCN) \cite{springenberg2014striving,long2015fully} as the generator to produce textures by sampling random latent variables as input. Unlike the conventional architecture of the generator in GAN~\cite{radford2015unsupervised,karras2019style}, we use convolutional operation in every layer, including the latent variable. Therefore, the latent variable is a tensor with spatial dimensions, which enables the generator to generate texture in multiple sizes as long as we expand the latent variable along the spatial dimensions. In the second stage, we search the best local pattern of the latent variable with a cropping technique\textemdash Toroidal Cropping (TC). After optimization, we can generate a large enough latent variable by tiling the local pattern. We input it to the FCN and finally get AdvTexture.

\begin{figure}[t]
\centering
\includegraphics[width=0.45\textwidth]{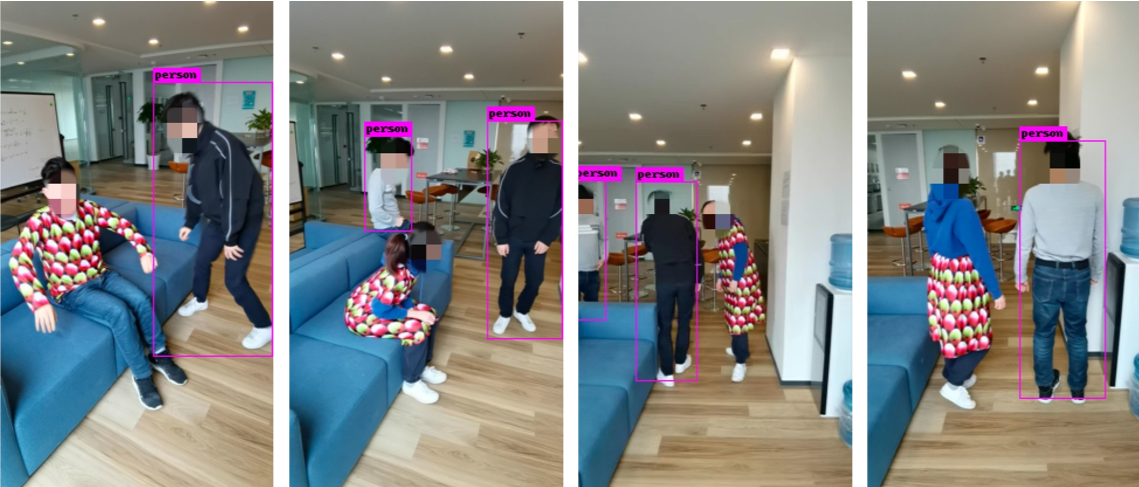}
\caption{Visualization of the adversarial effectiveness of AdvTexture when attacking YOLOv2. A dress, a T-shirt, and a skirt are tailored from a large polyester cloth material covered with the AdvTexture. The persons wearing the clothes failed to be detected by the detector.}
\label{fig:front}
\end{figure}

We implemented TC-EGA to attack various person detectors, and realized AdvTextures in the physical world. \cref{fig:front} shows some example attacks targeting YOLOv2. Our experiments showed that the clothes made from such textures significantly lowered the detection performance of different detectors.



\section{Related Work}
\label{sec:related}

Earlier works about adversarial examples \cite{szegedy2013intriguing,goodfellow2014explaining,kurakin2016adversarial} focused on digital attacks. Small adversarial noises can be added to the original images and make DNNs output wrong predictions, posing significant safety concerns to DNNs. 

Compared to digital adversarial attacks, physical adversarial attacks pose more risks in specific scenarios. Several methods~\cite{sharif2016accessorize,athalye2018synthesizing,brown2017adversarial,evtimov2017robust} have been proposed to attack image classification models physically. Sharif et al. \cite{sharif2016accessorize} designed a pair of glasses to attack face-recognition systems. Athalye et al. \cite{athalye2018synthesizing} generated robust 3D adversarial objects by introducing the Expectation over Transformation (EoT) \cite{athalye2018synthesizing} method. Brown et al. \cite{brown2017adversarial} deceived image classifiers by placing adversarial patches in the neighborhood of the objects. Evtimov et al. \cite{evtimov2017robust} misled road-sign classification by adhering black and white stickers to signs.

Recently, several methods \cite{thys2019fooling,thys2019fooling,huang2020universal,xu2020adversarial,wu2020making,wang2020can,hu2021naturalistic} were proposed to attack the DNN-based person detection systems. Thys et al. \cite{thys2019fooling} optimized an adversarial patch that can be attached to cardboard and held by a person. Huang et al.~\cite{huang2020universal} propose Universal Physical Camouflage Attack (UPC) to fool the detectors by simulating 3D objects in virtual environments. Xu et al.~\cite{xu2020adversarial} designed an adversarial T-shirt by introducing Thin Plate Spline (TPS)~\cite{bookstein1989principal,donato2002approximate} to simulate the deformation of clothes (e.g., wrinkles). Wu et al.~\cite{wu2020making} presented a systematic study of the attack on a range of detection models, different datasets, and objects. Wang et al.~\cite{wang2020can} masked the adversarial patch with preset logos and mapped it into 3D models. Hu et al.~\cite{hu2021naturalistic} used generative adversarial networks (GAN)~\cite{brock2019large,karras2019style} to craft more natural-looking adversarial patches.

Some works\cite{xu2020adversarial,huang2020universal,wang2020can} reported drops in the attack success rate when the viewing angles increased. According to Wang et al.~\cite{wang2020can}, part of the patches will not be captured when the camera rotates drastically. It can lead to underestimating the threat, whereas the cameras can be placed anywhere in real-world scenarios. 


\section{Methods}
\label{sec:AdvTexture}

\begin{figure*}[t]
\centering
\includegraphics[width=.7\textwidth]{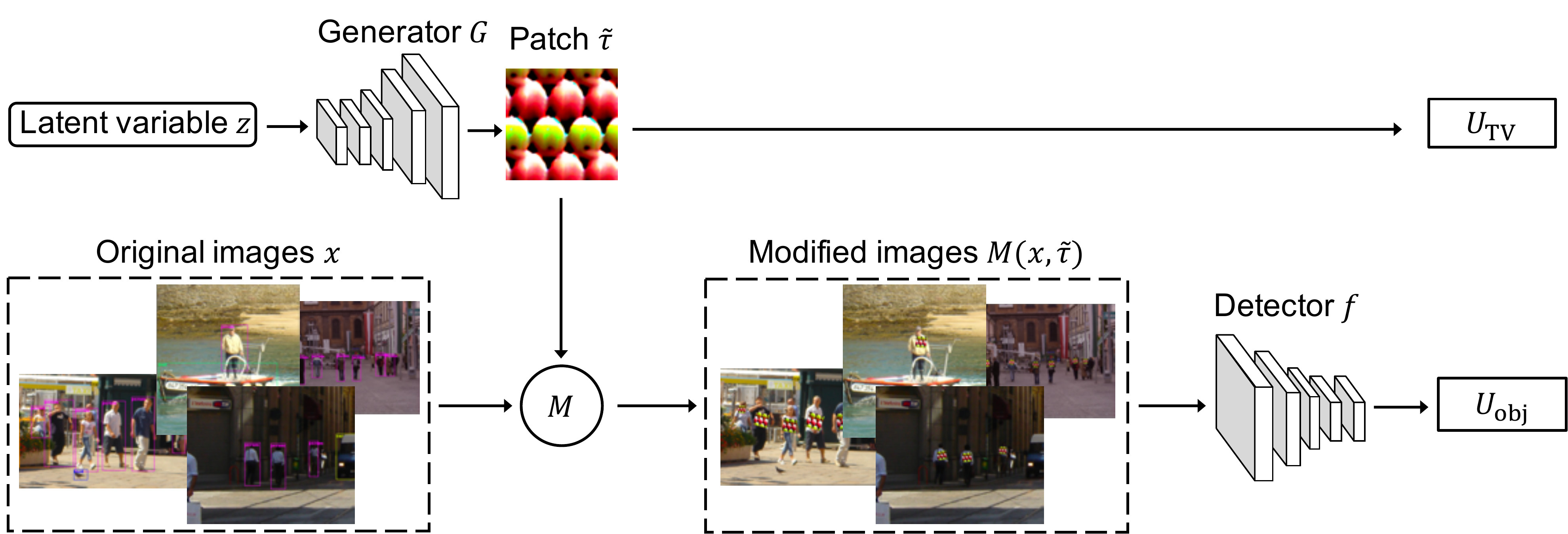}
\caption{The pipeline of the adversary objective function.}
\label{fig:pipline}
\end{figure*}

We aim to generate textures in arbitrary size, and when the textures are printed on cloth, any patch extracted from the cloth are effective in adversarial attack. We first introduce an adversarial patch generator and then describe TC-EGA based on the patch generator.

\subsection{Adversarial Patch Generator}
\label{sec:parameterize}


Let $\tau$ denote the whole cloth that is covered with AdvTexture, and $\Tilde{\tau}$ denote an extracted patch. We assume that $\Tilde{\tau}$ follows a distribution $p_{adv}$, such that the probability $p_{adv}(\Tilde{\tau})$ is higher when its adversarial effectiveness is more significant. We use an energy function $U(\Tilde{\tau})$ to model such a distribution:
\begin{equation}
  \label{eq:adv_dist}
  p_{adv}(\Tilde{\tau}) = \frac{e^{-U(\Tilde{\tau})}}{Z_U},
\end{equation}
where $Z_U=\int_{\Tilde{\tau}}{e^{-U(\Tilde{\tau})}}\mathrm{d}\Tilde{\tau}$ is called partition function. However, it is hard to sample from $p_{adv}(\Tilde{\tau})$ directly due to the partition function. Therefore, we use a parameterized generator $G_\varphi:z \to \Tilde{\tau}$ to approximate $p_{adv}(\Tilde{\tau})$, where $z\sim\mathcal{N}(0,I)$. We define $q_\varphi(\Tilde{\tau})$ as the distribution of $\Tilde{\tau}=G_\varphi(z)$, which can be written as
\begin{equation}
\label{eq:gen_dist}
q_\varphi(\Tilde{\tau})=\int \delta(\Tilde{\tau}-G_\varphi(z))p_z(z)\dif z,
\end{equation}
where $p_z$ is the probability density function (PDF) of the standard normal distribution $\mathcal{N}(0,I)$ and $\delta(\cdot)$ is the Dirac delta function. In order to represent $p_{adv}(\Tilde{\tau})$ more accurately, we tune $G_\varphi$ to minimize the KL divergence $\mathrm{KL}(q_\varphi(\Tilde{\tau})||p_{adv}(\Tilde{\tau}))$. With the aid of Deep InfoMax (DIM)~\cite{hjelm2018learning} we have the following theorem:
\begin{theorem}
\label{the:KL}
Minizing $\mathrm{KL}(q_\varphi(\Tilde{\tau})||p_{adv}(\Tilde{\tau}))$ is equivalent to
\begin{align}
&\min_{\varphi, \omega} \mathbb{E}_{\Tilde{\tau}\sim q_\varphi(\Tilde{\tau})}[U(\Tilde{\tau})] - I_{\varphi,\omega}^{\mathrm{JSD}}(\Tilde{\tau},z),
\label{eq:theo_loss}
\end{align}
where
\begin{align}
\mathcal{I}_{\varphi,\omega}^{\mathrm{JSD}}(\Tilde{\tau},z)=\ & \mathbb{E}_{(\Tilde{\tau}, z)\sim q_{\varphi}^{\Tilde{\tau}, z}(\Tilde{\tau}, z)}[-\mathrm{sp}(-T_{\omega}(\Tilde{\tau},z))]\notag\\
&- \mathbb{E}_{\Tilde{\tau}\sim q_{\varphi}(\Tilde{\tau}), z'\sim p_z(z')}[\mathrm{sp}(T_{\omega}(\Tilde{\tau},z'))],
\label{eq:info_loss}
\end{align}
$q_{\varphi}^{\Tilde{\tau}, z}$ denotes the joint distribution of $\Tilde{\tau}$ and $z$, and $\mathrm{sp}(t)=\mathrm{log}(1+e^t)$ is the softplus function. $T_\omega$ is a scalar function modeled by a neural network whose parameter $\omega$ must be optimized together with the parameter $\varphi$.
\end{theorem}
See \emph{Supplementary Materials} for the proof.


The objective function in \cref{eq:theo_loss} consists of two terms. The first term $\mathbb{E}_{\Tilde{\tau}\sim q_\varphi(\Tilde{\tau})}[U(\Tilde{\tau})]$ is called \emph{Adversary Objective Function} because minimizing it improves the the adversarial effectiveness of the generated patches. The second term $-\mathcal{I}_{\varphi,\omega}^{\mathrm{JSD}}(\Tilde{\tau},z)$ is called \emph{Information Objective Function} because minimizing it is equivalent to maximizing the mutual information of $z$ and $\Tilde{\tau}$~\cite{hjelm2018learning}, which requires different latent variables to generate different patches.

\subsubsection{The Adversary Objective Function}
\label{sec:adv_obj}

The adversary objective function $\mathbb{E}_{\Tilde{\tau}\sim q_\varphi(\Tilde{\tau})}[U(\Tilde{\tau})]$ can be estimated by sampling $z$ and generating $\Tilde{\tau}$:
\begin{equation}
 \frac{1}{N}\sum_{i=1}^{N}[U(G_\varphi(z_i))],
 \label{eq:adv_loss}
\end{equation}
where $\{z_i\}$ are the latent variables sampled from $\mathcal{N}(0,I)$, and $N$ denotes the total number of the samples. 


Now we need to set an appropriate energy function such that lowering the energy leads to detection failure of a person detector. We notice that detectors output multiple bounding boxes with a confidence score for each box when receiving an image. The boxes whose confidence scores are lower than a pre-specified threshold will then be filtered out. Therefore we choose the expectation of the confidence scores over boxes as a part of the energy function $U(\Tilde{\tau})$. Then minimizing the adv object function will lower the confidence scores of the boxes, which makes the boxes easily to be filtered out.

Specifically, we randomly generate patches in every step, and apply a set of physical transformations such as randomizing the scales, contrast, brightness and additional noise according to Expectation over Transformation (EoT)~\cite{sharif2016accessorize,thys2019fooling}. We also incorporate random Thin Plate Spin (TPS)~\cite{xu2020adversarial,donato2002approximate} deformation as an additional random transformation. We then attach the patches randomly to the persons according to the predicted boxes on the images $x$ from the training set. We use $M(x,\Tilde{\tau})$ to denote the above process, and obtain the modified images which are then be sent into the target detector. This part of the energy function is thus defined as
\begin{equation}
   U_{\mathrm{obj}} = \mathbb{E}_{x, M}[f(M(x, \Tilde{\tau}))],
\end{equation}
where $f$ denotes confidence scores of the boxes predicted by the target detector.

We use a differentiable variation of total variance (TV) loss~\cite{sharif2016accessorize} as another part of the energy function to encourage the patches to be smoother:
\begin{equation}
   U_{\mathrm{TV}} = \sum_{i,j}{|\tau_{i,j} - \tau_{i+1,j}|+|\tau_{i,j} - \tau_{i,j+1}|}
\end{equation}
Together, we form the energy function as 
\begin{equation}
  U(\Tilde{\tau}) = \frac{1}{\beta}(U_{\mathrm{obj}} + \alpha U_{\mathrm{TV}}),
\end{equation}
where $\alpha$ and $\beta$ are coefficients. See \cref{fig:pipline} for the illustration. When minimizing the adversary objective function, each part of the energy function will be minimized together.

\subsubsection{The Information Objective function}
\label{sec:info_obj}
\begin{figure}[t]
\centering
\includegraphics[width=0.7\linewidth]{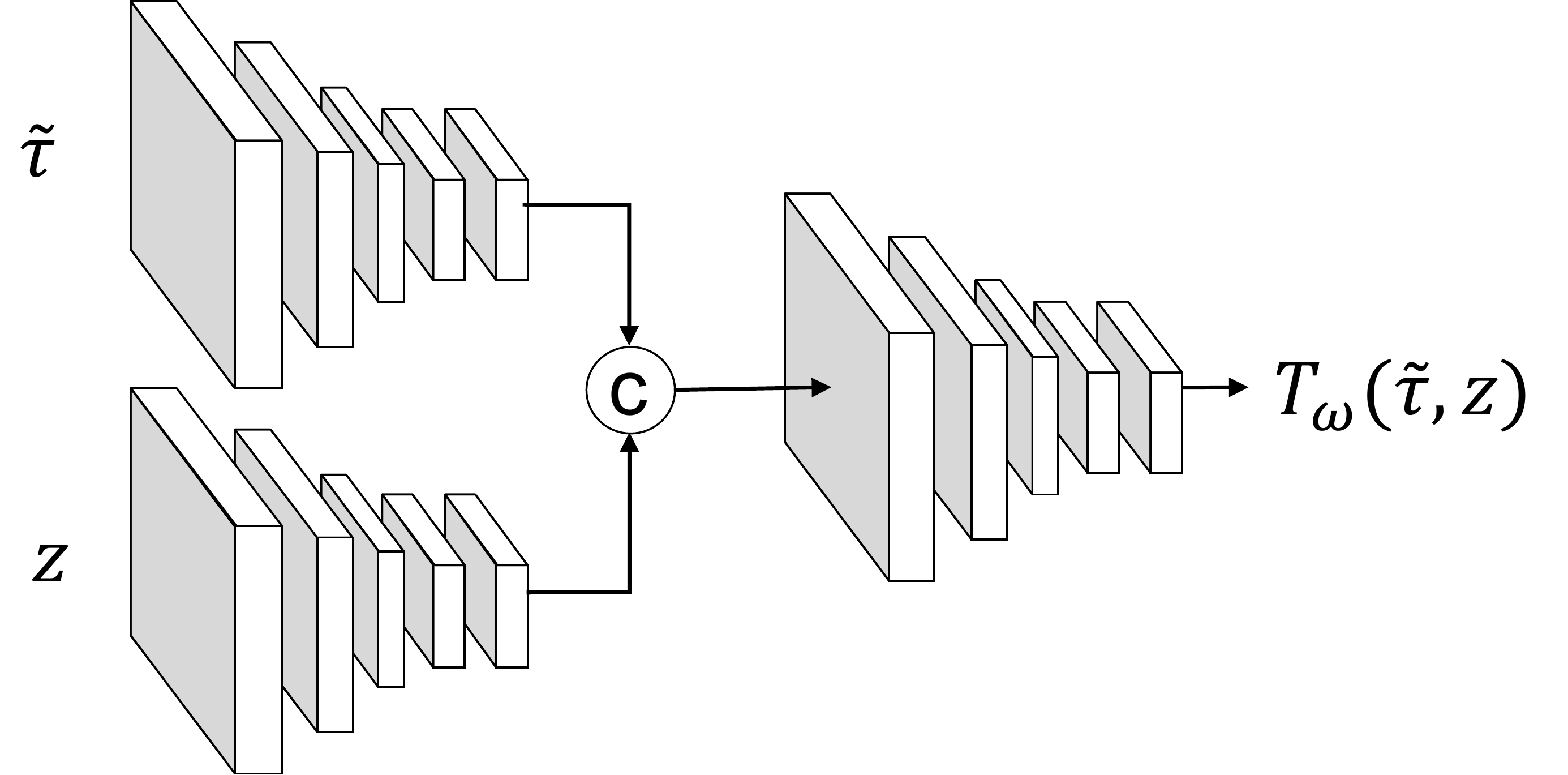}
\caption{The architecture of the auxiliary network $T_\omega$. It has two inputs, $\Tilde{\tau}$ and $z$, and outputs a scalar value $T_{\omega}(\Tilde{\tau},z)$. The operation \emph{c} in the figure stands for concatenation.}
\label{fig:aux}
\end{figure}

As described in \cref{eq:info_loss}, we use an auxiliary network $T_\omega$ to increase the mutual information of $z$ and $\Tilde{\tau}$. We illustrate the architecture of $T_\omega$ in \cref{fig:aux}. \cref{eq:info_loss} has two terms, and estimating each of them needs random sampling. Following the previous work~\cite{hjelm2018learning}, to estimate the first term, we first sample $z$ from $\mathcal{N}(0, I)$, and then generated $\Tilde{\tau}$ by $G_\varphi(z)$ in each training step. To estimate the second term, we keep $\Tilde{\tau}$ and resample $z$. 


During training, we minimize the adversarial objective function and the information objective function simultaneously. Therefore, the distribution $q_\varphi$ can approximate to $p_\mathrm{adv}$, which means the the generated patches $\Tilde{\tau}$ can be adversarial to the target detector.

\subsection{Toroidal-Cropping-based Expandable Generative Attack}

In \cref{sec:parameterize}, we have described the method to train a generator for adversarial patches $\Tilde{\tau}$. In this section, we used TC-EGA to generate AdvTextures $\tau$ based on the adversarial patch generator. We leverage a specific network architecture and a sample technique to extend adversarial patches to adversarial textures. TC-EGA has two stages. In the first stage, we train a fully convolutional network (FCN) \cite{springenberg2014striving,long2015fully} to help sample from the distribution of adversarial textures. In the second stage, we search the best latent representation to yield the most effective adversarial texture.

\subsubsection{Stage One: Train an Expandable Generator}
\label{sec:stage1}

\begin{figure}[t]
\centering
\begin{subfigure}{1.05\linewidth}
\includegraphics[width=\textwidth]{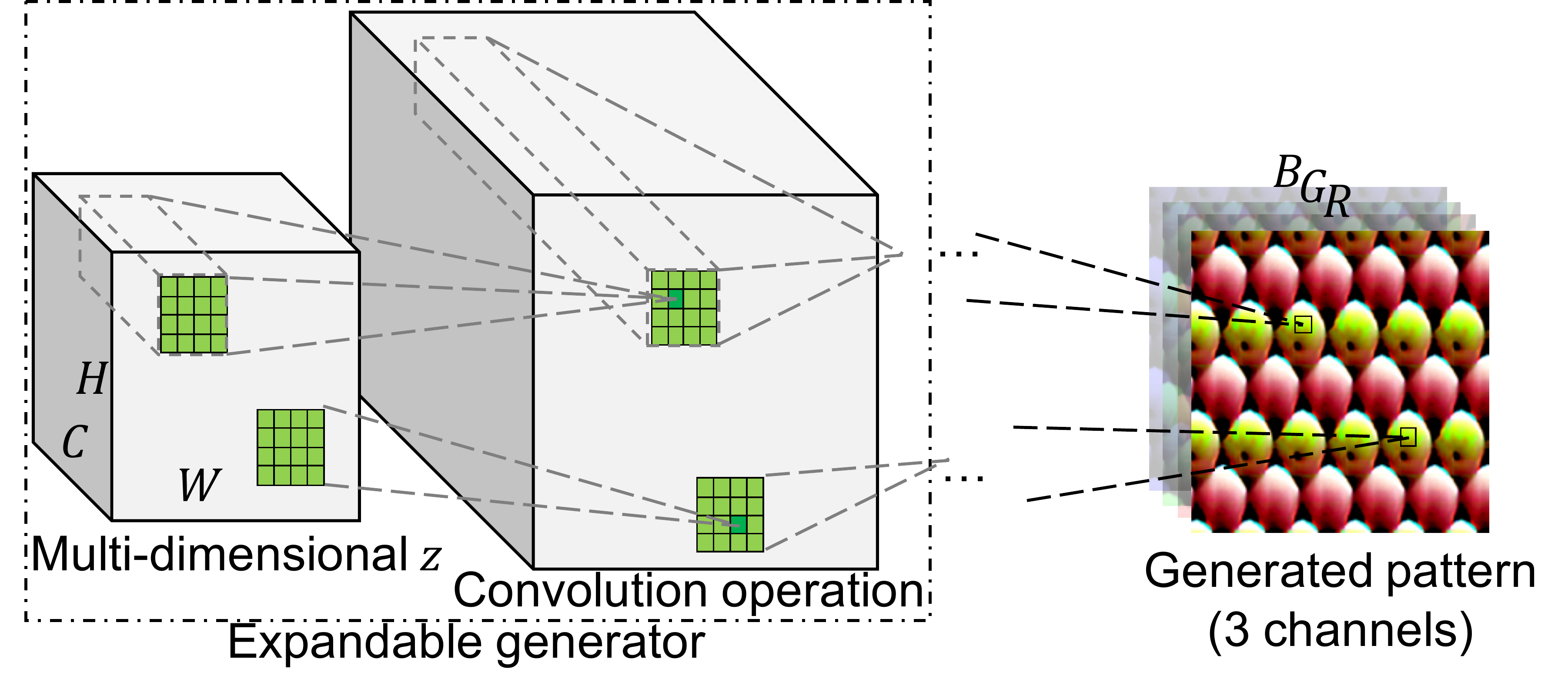}
\caption{}
\label{fig:fcn}
\end{subfigure}
\begin{subfigure}{0.9\linewidth}
\includegraphics[width=\textwidth]{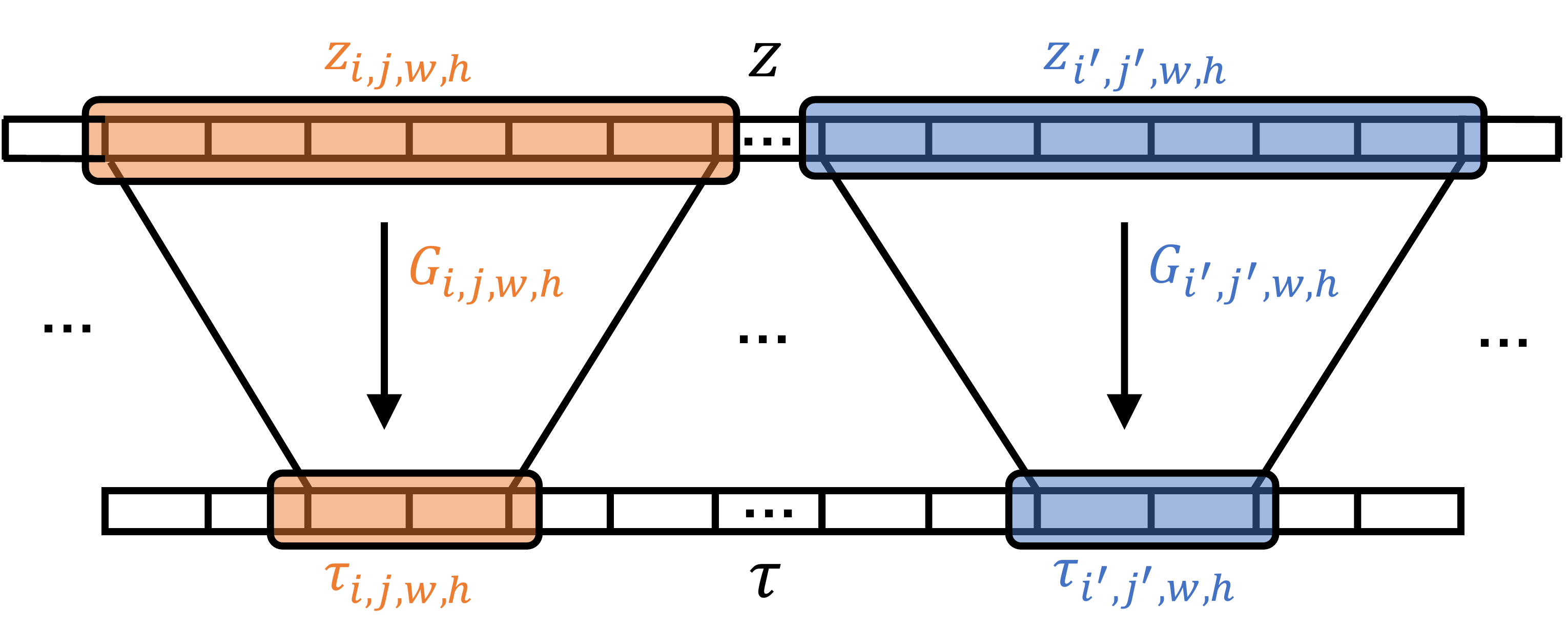}
\caption{}
\label{fig:fcn_s}
\end{subfigure}
\caption{(a) Illustration of the FCN generator. All layers of the generator network are convolutional layers with zero padding, including the first layer. (b) Each patch $\tau_{i,j,w,h}$ extracted from position $i,j$ can be regarded as the output of a sub-generator $G_{i,j,w,h}$ when the input is $z_{i,j,w,h}$.}
\end{figure}

We aim to train a generator so that it can generate patches in arbitrary size easily by taking a random $z$ as input. The critical point is to endow the generator with translation invariant property by constructing an FCN, where all layers are convolutional layers with zero padding, including the first layer that inputs the latent variable (See \cref{fig:fcn}). The latent variable is a $B \times C \times H \times W$ tensor where $B$ is the batch size, $C$ is the number of channels, and $H$, $W$ are height and width, respectively.

Here we show the reason for using FCN. We assume that the overall texture $\tau$ is generated by a global generator $G:z \to \tau$ with hidden variable $z\sim \mathcal{N}(0,I)$. We denote the extracted patch by $\tau_{i,j,w,h}$ whose center is located at the position $(i,j)$ of the overall texture and has a shape of $(w,h)$. Moreover, the patch $\tau_{i,j,w,h}$ can be regarded as the output of a sub-generator $G_{i,j,w,h}:z_{i_z,j_z,w_z,h_z}\to \tau_{i,j,w,h}$, where $z_{i_z,j_z,w_z,h_z}$ is the component of $z$ that consists of all the elements dependent to $\tau_{i,j,w,h}$ (see \cref{fig:fcn_s}). Assuming that $\tau_{i,j,w,h}$ follows a distribution $\mathcal{T}_{i,j,w,h}$. We have the following theorem and corollary.

\begin{theorem}
\label{theo:irrelevant}
Let $\tau_1 =G_1(z_1)$, $\tau_2 =G_2(z_2)$, $z_{1}\sim\mathcal{Z}_{1}$, $z_{2}\sim\mathcal{Z}_{2}$, $\tau_{1}\sim\mathcal{T}_{1}$, $\tau_{2}\sim\mathcal{T}_{2}$. If $\mathcal{Z}_{1}$ is identical to $\mathcal{Z}_{2}$ and $G_{1}$ is equivalent to $G_{2}$, then $\mathcal{T}_{1}$ is identical to $\mathcal{T}_{2}$.\\
\end{theorem}
\begin{corollary}
\label{theo:fcnn}
$G_{i,j,w,h}$ and $\mathcal{T}_{i,j,w,h}$ are irrelevant to $i,j$, i.e., $G_{i,j,w,h}=G_{w,h}$ and $\mathcal{T}_{i,j,w,h}=\mathcal{T}_{w,h}$, if $G$ is an FCN and the input $z \sim \mathcal{N}(0, I)$.
\end{corollary}
See \emph{Supplementary Materials} for the proofs. Therefore, as long as the sub-generator $G_{w,h}$ is trained to approximate the distribution of $\mathcal{T}_{w,h}$ to $p_{\mathrm{adv}}$, any patch extracted from the overall texture with shape $(w,h)$ also approximately follows $p_{\mathrm{adv}}$, i.e., it has adversarial effectiveness. Moreover, due to the translation invariant property of the convolutional operation, the sub-generator $G_{w,h}$ and the global generator can share the same architecture and parameters except for the different spatial shape $H$ and $W$ of the latent variable $z$. As a result, we only need to train a small generator.

Note that the height $H$ and width $W$ of the hidden variable $z$ can not be too small, otherwise the output will be too small to crop a patch in spatial shape $(w,h)$. We denote the minimum spatial sizes by $H_{\mathrm{min}}$ and $W_{\mathrm{min}}$. During training, we sampled a small $z$ in shape $B \times C \times H_{\mathrm{min}} \times W_{\mathrm{min}}$ and generated the corresponding patches in each training step. After that, we can produce different textures of arbitrary sizes by randomizing $z$ with any $H\geq H_\mathrm{min}$ and $W\geq W_\mathrm{min}$.

\subsubsection{Stage Two: Find the Best Latent Pattern}
\label{sec:stage2}

After training, the generator can generate different textures by sampling latent variables. In order to find the best texture for adversarial attacks, we propose to go one step further, that is, to optimize the latent variable with the parameters of the generator frozen. However, since the texture has no specific shape and the size of the latent variable needs to be large enough to produce a large textured cloth, directly optimizing the latent variable is difficult.

\begin{figure}[t]
\centering

\begin{subfigure}{0.5\textwidth}
\includegraphics[width=\textwidth]{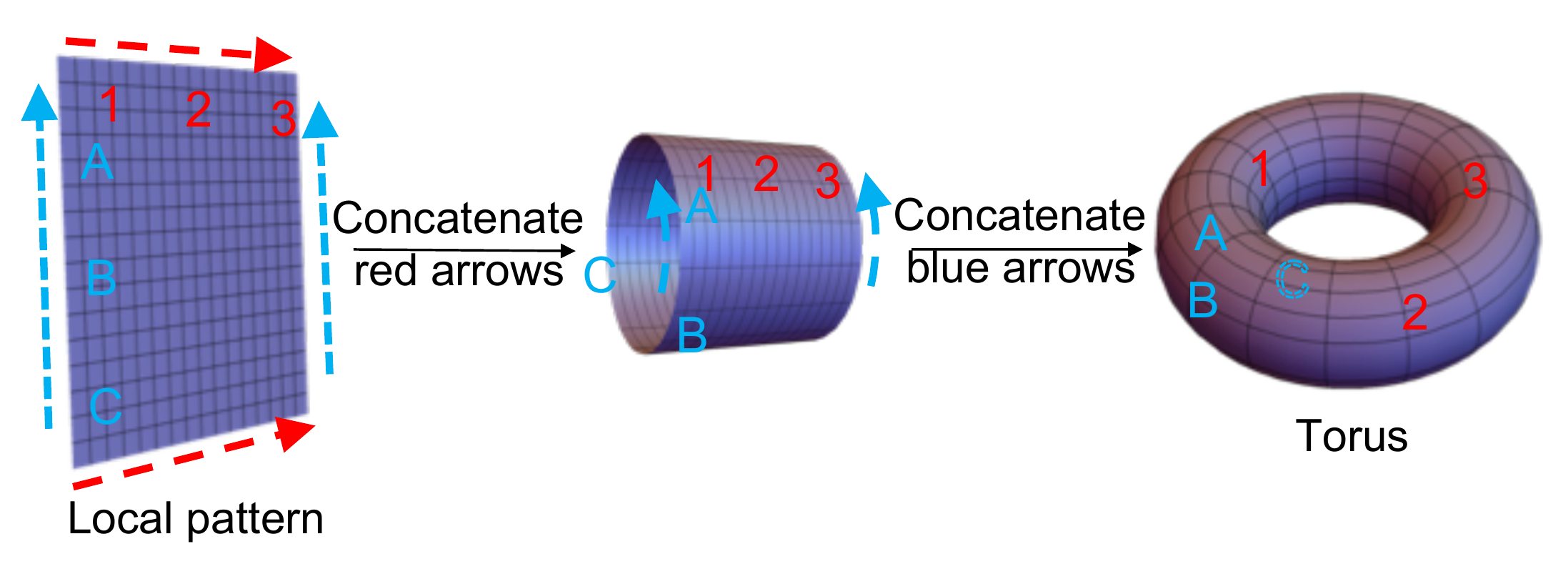}
\caption{}
\label{fig:localTCA}
\end{subfigure}
\begin{subfigure}{0.2\textwidth}
\includegraphics[height=.95in]{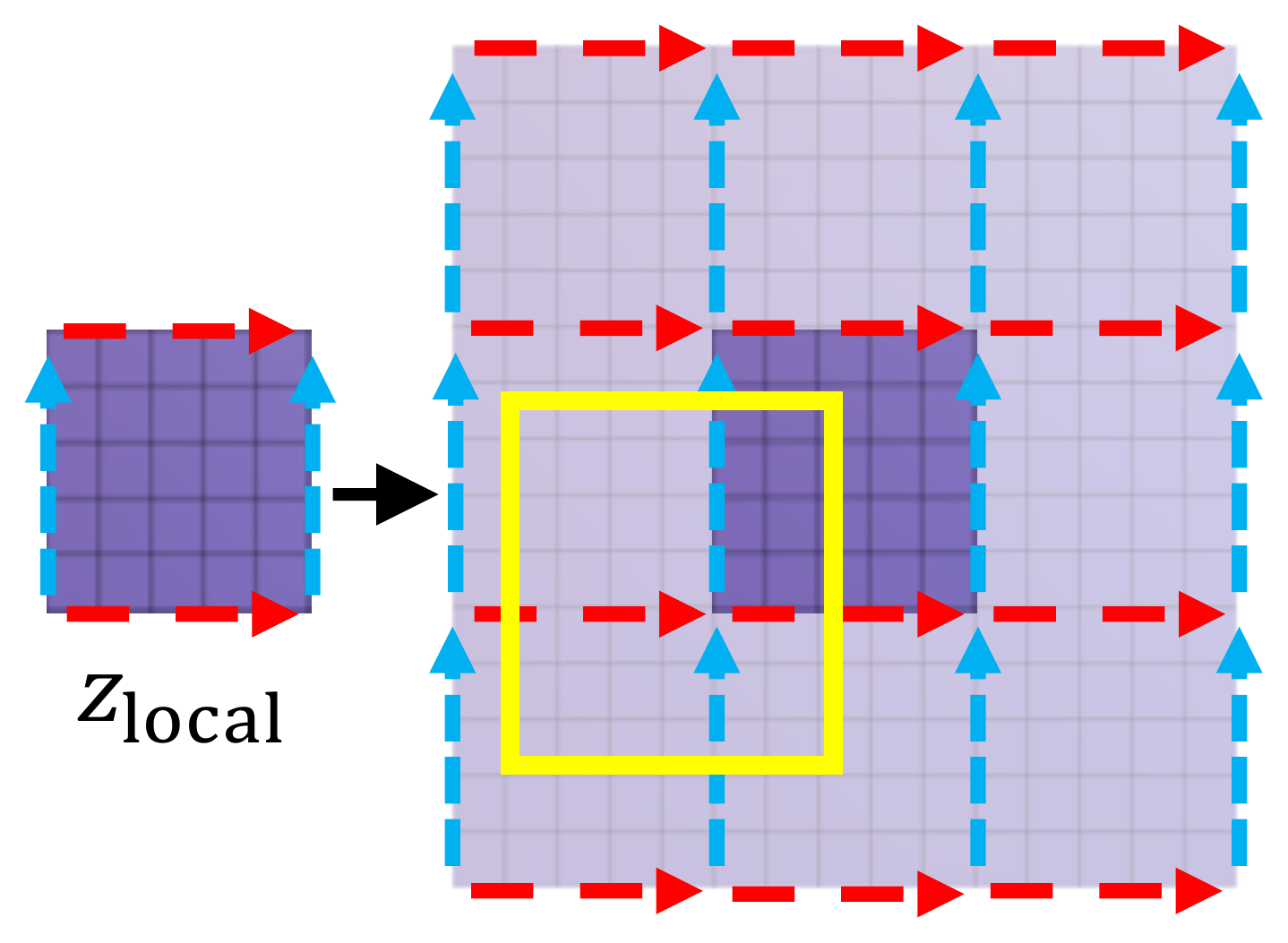}
\caption{}
\label{fig:zTCA}
\end{subfigure}
\begin{subfigure}{0.27\textwidth}
\includegraphics[height=.95in]{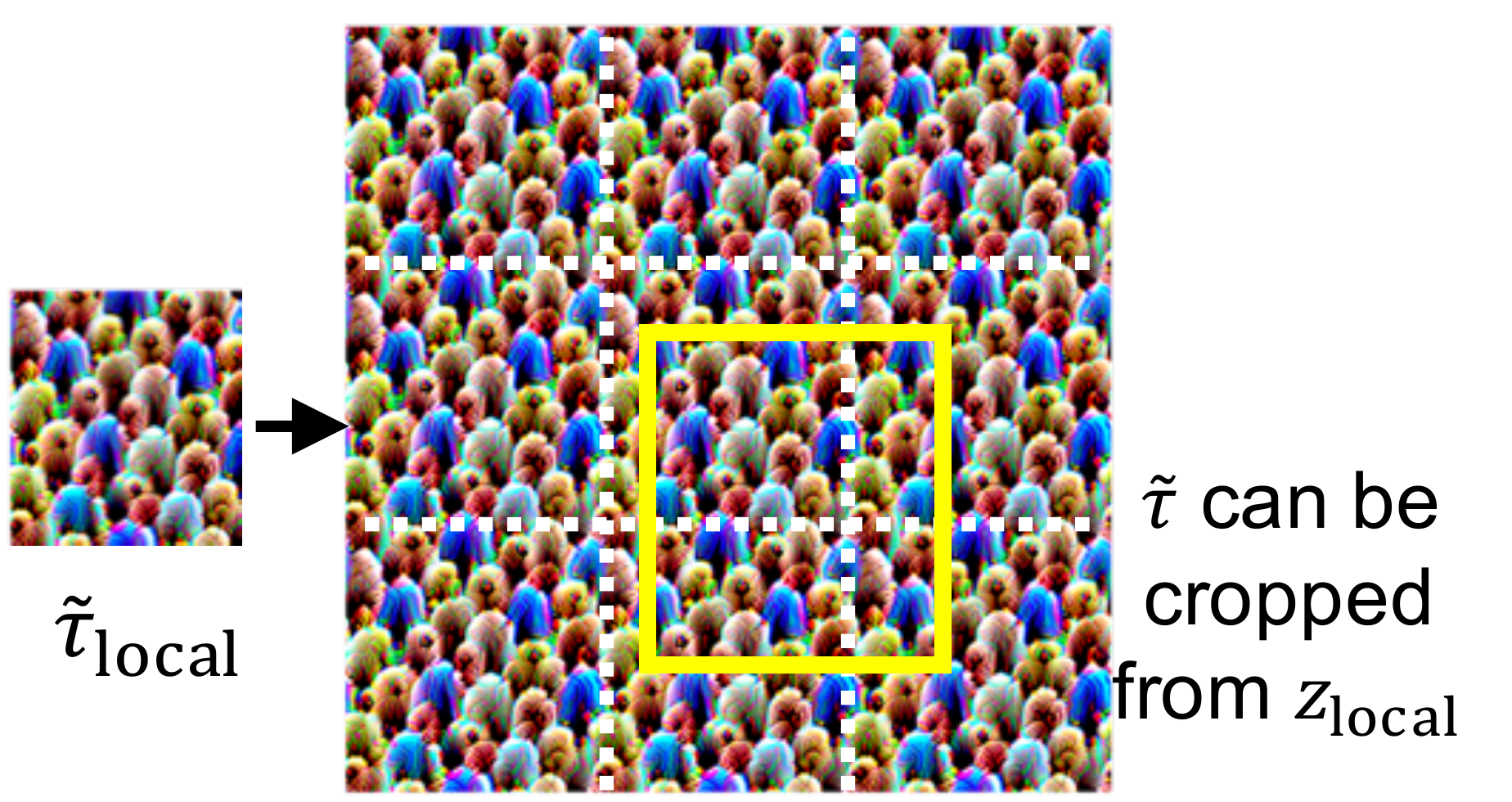}
\caption{}
\label{fig:tTCA}
\end{subfigure}
\caption{Illustration of Toroidal Cropping. (a) By first concatenating its horizontal edges (red arrow) and then concatenating the vertical edges (blue arrow), the local pattern can be folded to a torus. (b) The latent variable in arbitrary shape can be created by tiling the local pattern side by side, thus the variable cropped at the junctions is equivalent to that cropped on the torus, meaning the pattern is still continuous. (c) This cropping technique also applies to the pixel space. See \cref{sec:basemodel} for this variant.}
\label{fig:TCA}
\end{figure}

Inspired by the unfolding of torus in topology which supports up-down and left-right continuation \cite{hatcher2002algebraic} (\cref{fig:localTCA}), we introduce the Toroidal Cropping (TC) technique, which aims to optimize a local pattern $z_{\mathrm{local}}$ as a unit such that the final latent variable $z$ can be produced by tiling multiple identical units. In detail, $z_{\mathrm{local}}$ can be parameterized as a tensor in shape $B \times C \times L \times L$ with a shape hyper-parameter $L$, which can be regarded as the unfolded plane of a two-dimensional torus $\mathbb{T}^2$ in topology (\cref{fig:localTCA}). Therefore the latent variable in arbitrary shape can be cropped from $z_\mathrm{local}$ in a recursive manner (\cref{fig:zTCA}), which can be regarded as cropping on the torus. We denote such crop operation by $\mathrm{Crop}_{\mathrm{torus}}$.

During optimization, we randomly sample the latent variables $z_{\mathrm{sample}}$ in shape $B \times C \times H_{\mathrm{min}} \times W_{\mathrm{min}}$ by such cropping technique. Since we only consider the adversarial effectiveness in this stage, we generated patches by $z_{\mathrm{sample}}$ and minimized the adversary loss (\cref{eq:adv_loss}). After optimization, one can produce a latent variable with arbitrary size by tiling $z_{\mathrm{local}}$.

\section{Experiment settings}
\label{sec:experiment}
\subsection{Subjects}
We recruited three subjects (mean age: $24.0$; range: $21-26$; two males and one female) to collect physical test set. The recruitment and study procedures were approved by the Department of Psychology Ethics Committee, Tsinghua University, Beijing, China.

\subsection{Dataset}
We employed the Inria Person dataset \cite{dalal2005histograms} as our training set. It is a dataset for pedestrian detection, which consists of $614$ images for training and $288$ for testing. We evaluated the patch-based attack on the Inria test set. For physical evaluation, we produced clothes covered with different adversarial textures. Three subjects wore different adversarial clothes and turned a circle slowly in front of a camera which was fixed at $1.38$ meters above the ground. The distance between the camera and person is fixed to $2\;\mathrm{m}$ unless otherwise specified. We recorded two videos for each subject and each adversarial piece of clothing. One of the video was recorded indoor (lab room), and the other was recorded in outdoor (brick walkway). We then extracted $32$ frames from each video. We recorded $3 \times 2=6$ videos and collected $6 \times 32 = 192$ frames for each adversarial piece of clothing. we labeled them manually to construct a test set.


\subsection{Baseline Methods}
\label{sec:basemodel}

We evaluated the adversarial patches produced by Thys et al.~\cite{thys2019fooling} and Xu et al. \cite{xu2020adversarial}, and named them by \emph{AdvPatch} and \emph{AdvTshirt}, respectively. We copied the patterns from their original papers. We also tiled AdvPatch and AdvTshirt to form textures with repeated patterns. These two variants are called \emph{AdvPatchTile} and \emph{AdvTshirtTile}. In addition, we evaluated a texture with repetitive random colors, which is denoted by \emph{Random}


Moreover, TC-EGA has multiple components and some of them could be applied separately to craft adversarial textures. To investigate the performance of each component, we designed three variants of TC-EGA, as described below. 

\paragraph{Expandable Generative Attack (EGA)} We trained an FCN as the first stage of TC-EGA without optimizing the best latent variable. During evaluation, the final texture can be generated by a latent variable in arbitrary size and sampled from a standard normal distribution.

\paragraph{Toroidal Cropping Attack (TCA)} We directly optimized the texture instead of training an FCN to generate texture. Specifically, we initialized a local texture pattern of $300\times300$ pixels, and randomly extracted a patch by size $150\times150$ from the texture by Toroidal Cropping in each optimization step.

\paragraph{Random Cropping Attack (RCA)} We directly optimized a large patch whose size is fixed. We initialized the large patch and randomly cropped a small patch by size $150\times150$ during optimization. This method is named Random Cropping Attack (RCA). We implemented two attacks, RCA$2\times$ and RCA$6\times$, where the sizes of the large patches are $300\times300$ and $900\times900$, respectively.

\subsection{Implementation Details}

We crafted AdvTexture to mainly fool YOLOv2~\cite{redmon2017yolo9000}, YOLOv3~\cite{redmon2018yolov3}, Faster R-CNN~\cite{ren2016faster} and Mask R-CNN~\cite{he2017mask}. The detectors were pre-trained on MS COCO dataset~\cite{lin2014microsoft}. Their outputs were filtered to output the person class only.

For each target detector, we first extracted the predicted bounding boxes on the images from the training set with a Non-Maximum Suppression (NMS) threshold $0.4$. We chose the boxes whose confidence was larger than a certain threshold ($0.5$ for YOLOv2 and YOLOv3, and $0.75$ for Faster and Mask R-CNN). We additionally filtered out boxes with areas smaller than $0.16\%$ of the entire images for Faster and Mask R-CNN. Then, as we described in \cref{sec:adv_obj}, we attached the extracted patches to the persons and input the modified images to the detector during optimization.

Moreover, we applied the Adam \cite{kingma2014adam} optimizer to optimize parameters in both stages. The hyper-parameters are listed as follows. (1) Stage one: The initial learning rate to train the generator was $0.001$. The generator was a $7$-layer FCN whose input was the latent variable $z$ with size $B\times128\times9\times9$. The size of the corresponding output was $B\times3\times324\times324$, where the second dimension stands for the RGB channels. (2) Stage two: We optimized a local latent variable $z_{\mathrm{local}}$ with size $1\times128\times4\times4$, followed by the Toroidal Cropping technique to produce samples of $z$ with size $B\times128\times9\times9$. The learning rate of the optimization was $0.03$. 

To physically implement AdvTexture, we printed the texture on a polyester cloth material by digital textile printing. Afterwards, we hired a professional tailor to produce adversarial clothes including T-shirts, skirts and dresses.

\section{Results}
\cref{fig:pattern_vis} shows some textures obtained by different methods, and more can be found in \emph{Supplementary Materials}.
\subsection{Patch-Based Attack in the Digital World}
\label{sec:patch}
We first evaluated the attacks in the form of the patch-based attack in the digital world. Specifically, we randomly extracted patches from the textures when evaluating most methods except for AdvPatch and AdvTexture. We denote such patches by \emph{resampled} patches. We then attached the patches to the images from the Inria test set the same way as crafting the adversarial patches. We used the bounding boxes proposed by the target detectors on the original test images with a confidence threshold of $0.5$ as the ground truth. We computed the average precision (AP) of the proposed bounding boxes on the modified test images to measure the adversarial effectiveness. Note that lower AP indicates stronger attack.



\begin{figure}[t]
   \centering
   \begin{subfigure}{0.12\textwidth}
   \includegraphics[width=.9\textwidth]{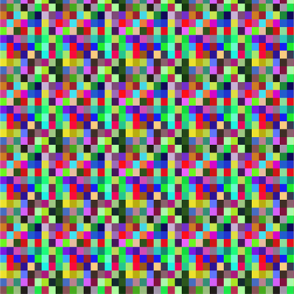}
   \caption{Random}
   \label{fig:Random}
   \end{subfigure}
   \begin{subfigure}{0.12\textwidth}
   \includegraphics[width=.9\textwidth]{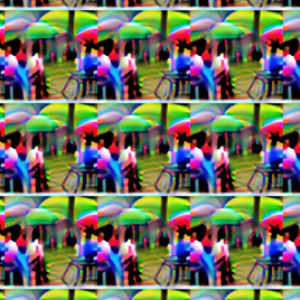}
   \caption{AdvPatchTile}
   \label{fig:AdvPatchTile}
   \end{subfigure}
   \begin{subfigure}{0.12\textwidth}
   \includegraphics[width=.9\textwidth]{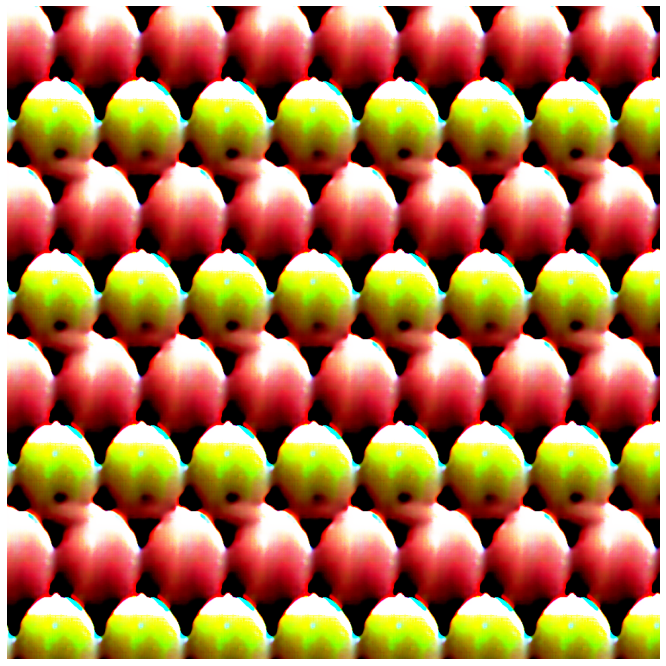}
   \caption{TC-EGA}
   \label{fig:v2}
   \end{subfigure}
   \caption{Visualization of different textures. (a) The texture with repetitive random colors. (b) The texture formed by tiling an adversarial patch~\cite{thys2019fooling} repeatedly. (c) The texture produced by TC-EGA to attack YOLOv2.}
   \label{fig:pattern_vis}
\end{figure}

\begin{table}
\centering
\small
\begin{threeparttable}
\begin{tabular}{lccc}
\toprule
Method                  & AP      & Expandable   & Resampled   \\ \midrule
Clean                   & $1.000$   & &              \\
Random                  & $0.963$ & \ding{51}&\ding{51}             \\
AdvPatch~\cite{thys2019fooling}                & $\textbf{0.352}$ &\ding{55}&\ding{55}     \\
AdvPatchTile            & $0.827$ &\ding{51}&\ding{51}     \\
AdvTshirt~\cite{xu2020adversarial}               & $0.744$\tnote{*} &\ding{55}&\ding{55}     \\
AdvTshirtTile           & $0.844$ &\ding{51}&\ding{51}     \\
 \midrule
TC-EGA                  & $\textbf{0.362}$ &\ding{51}&\ding{51}     \\
EGA                     & $0.470$ &\ding{51}&\ding{51}     \\
TCA                     & $0.664$ &\ding{51}&\ding{51}     \\
RCA$2\times$            & $0.606$ &\ding{55}&\ding{51}     \\
RCA$6\times$            & $0.855$ &\ding{55}&\ding{51}     \\
\bottomrule
\end{tabular}
\end{threeparttable}
\captionsetup{width=\linewidth}
\captionof{table}{The APs of YOLOv2 under different attacks on Inria test set. \emph{Expandable} denotes whether the methods can produce textures in arbitrary size. \emph{Resampled} denotes whether the patches are randomly extracted.}
\label{tab:dig_main}
\end{table}

\begin{figure}[tbp]
    \centering
    \includegraphics[width=0.5\textwidth]{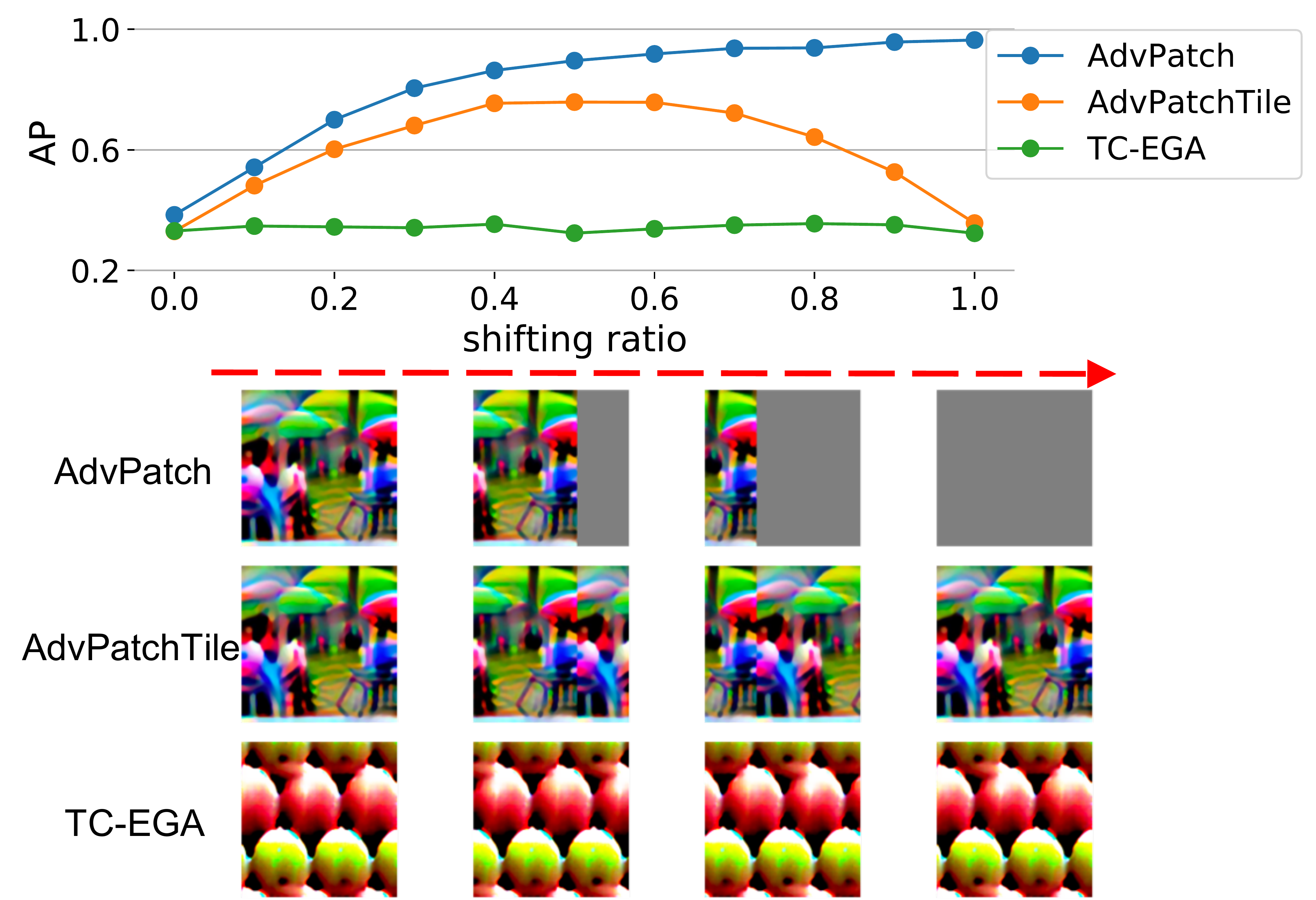}
   \caption{Numerical study of the segment-missing problem. The patches are cropped near the original patches with a shifting ratio. For AdvPatch as an example, the shifting ratio is $0.0$ when the cropped patch is precisely the original patch. The shifting ratio is $1.0$ when the original patch is shifted totally outside the cropping range.}
    \label{fig:shift}
\end{figure}



\cref{tab:dig_main} presents the AP of YOLOv2 in different conditions. \emph{clean} denotes the AP on the original test set. Since we used the detector's prediction on the original images as the ground truth, the AP is $1.000$. The AdvPatch lowered the AP of YOLOv2 to $0.352$\footnote{We reproduced an adversarial patch according to their released code https://gitlab.com/EAVISE/adversarial-yolo. The reproduced patch got an AP of $0.378$. We used the patch that is copied from their paper in all the experiments.}. 

Compared to AdvPatch, the expandable variant AdvPatchTile increases the AP from $0.352$ to $0.827$. Since AdvTshirt was trained on a different dataset (its authors' private dataset), it only got an AP of $0.744$. Similarly, AdvTshirtTile increases the AP to $0.844$. We attribute the increase to the segment-missing problem. Compared to its variants, TC-EGA got the lowest AP $0.362$, which was also the lowest among all the resampled patches. AdvPatch made the AP slightly lower than TC-EGA. However, it is not expandable and thus unsuitable for the attack at multiple viewing angles. Moreover, EGA decreased the AP to $0.470$, TCA created expandable patches with AP $0.664$. It was lower than AdvPatchTile, which indicates the effectiveness of the Toroidal Cropping technique. Moreover, RCA$6\times$ was much worse than RCA$2\times$, which indicates difficulties in optimizing a large patch.


We further investigated the segment-missing problem by evaluating the adversarial effectiveness of the patches which is cropped at shifted positions (See \cref{fig:shift}). The patch-based attack, AdvPatch, became less effective when the shifting ratio increased. Tiling the patches alleviated the problem, but was still problematic. The texture generated by TC-EGA was robust during shifting.

The results of other detectors attacked by TC-EGA in the digital world are shown in \emph{Supplementary Materials}.

\subsection{Attack in the Physical World}
\label{sec:physical}

\cref{fig:clothes} shows the produced clothes by different methods, and more can be found in \emph{Supplementary Materials}. 


\begin{figure}[t]
\centering
\includegraphics[width=0.5\textwidth]{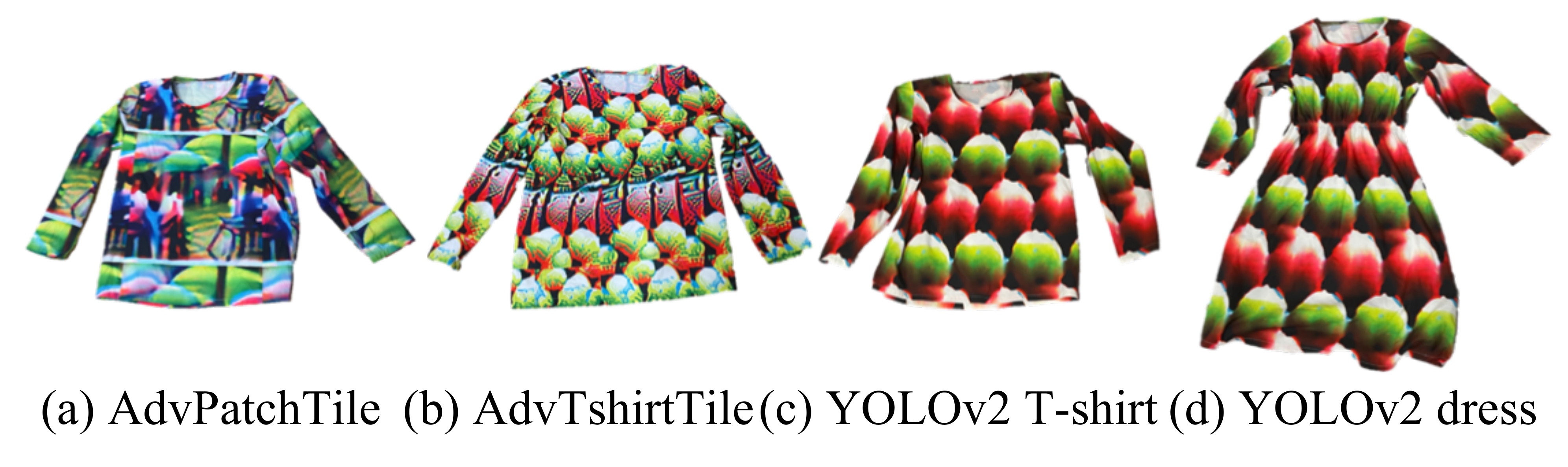}
\caption{Real-world adversarial clothes.}
\label{fig:clothes}
\end{figure}

\begin{figure}[tbp]
    \centering
    \includegraphics[width=.5\textwidth]{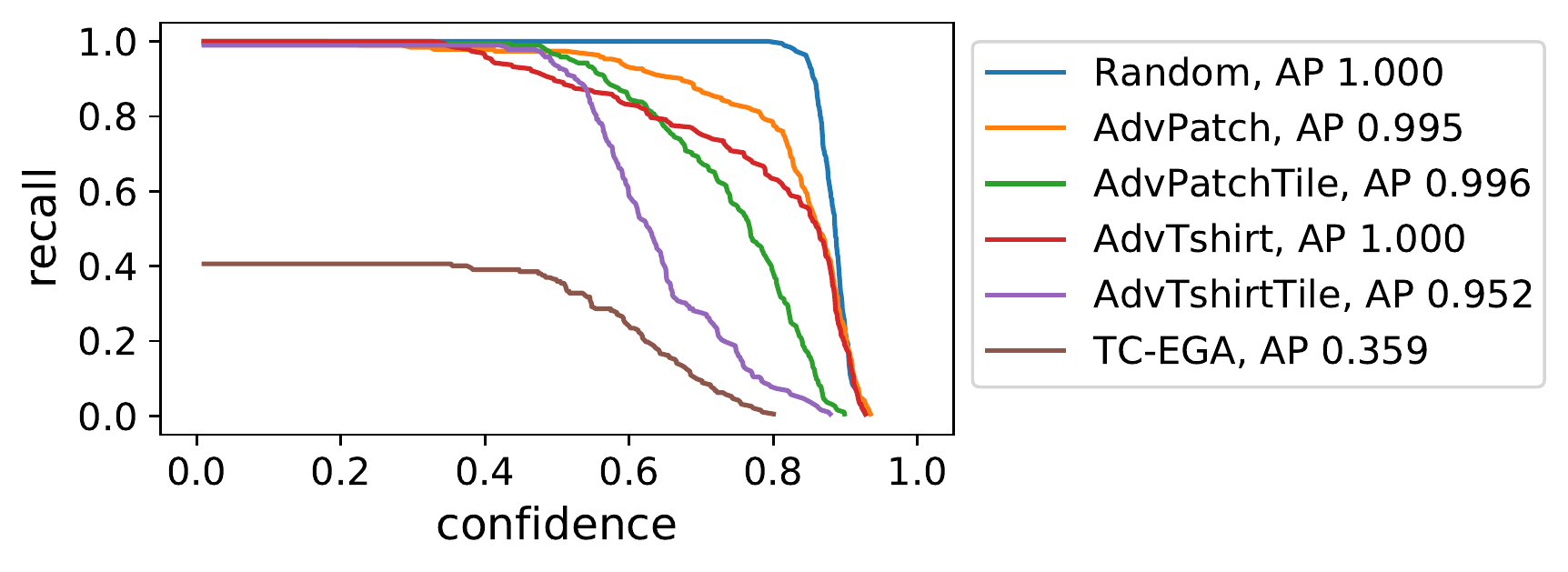}
    \caption{The recall v.s confidence curves and APs on the physical adversarial test set. The target network is YOLOv2.}
    \label{fig:phy_v2}
\end{figure}

We first compared different methods on YOLOv2. Since the boxes predicted by the detectors can be filtered by a particular confidence threshold, we plotted the recall-confidence curve in \cref{fig:phy_v2} and showed their APs in the legend. Remember that \emph{recall} denotes the fraction of the boxes that are successfully retrieved. These boxes are filtered by a confidence threshold. Therefore, for each particular confidence threshold, lower recall denotes better adversarial effectiveness. From \cref{fig:phy_v2}, the tiled variants of both AdvPatch and AdvTshirt were more effective than the original method. TC-EGA outperformed among all the methods by the lowest recall-confidence curve and the lowest AP.

Moreover, we used another metric to evaluate the effectiveness of the attacks. Specifically, for each input image we collected the target detector's predicted bounding boxes whose confidence score is larger than a certain confidence threshold. As long as one of these boxes has an Intersection over Union (IoU) with the ground-truth box greater than $0.5$, the detector is considered to have correctly detected. We defined Attack Success Rate (ASR) as the fraction of the test images that are not correctly predicted. Since the ASR is relevant to the confidence threshold, we calculated the mean value of the ASR, namely mASR, under multiple thresholds. The thresholds were $0.1, 0.2,..., 0.9$ in our experiment.

\begin{figure}[t]
    \centering
    \includegraphics[width=.5\textwidth]{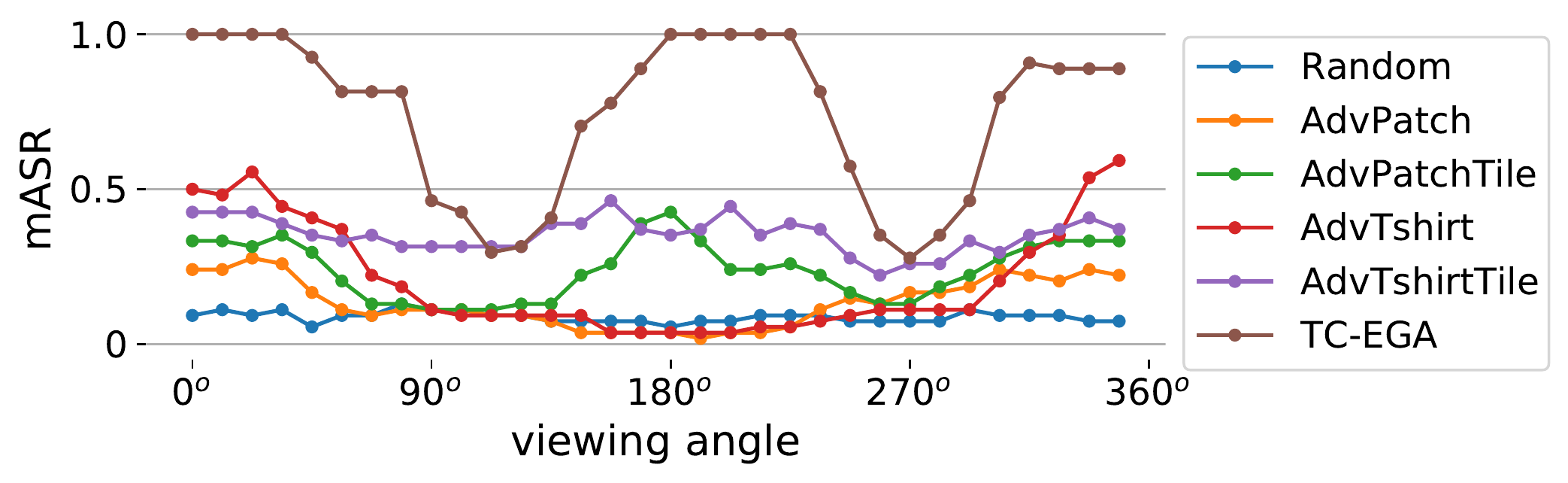}
    \caption{The mASRs of the attacks at multiple viewing angles.}
    \label{fig:phy_dis}
\end{figure}

\cref{fig:phy_dis} presents the mASRs in multiple viewing angles. Compared to the random texture, AdvPatch and AdvTshirt was effective when the persons faced the camera (the viewing angle is $0^{\circ}$ or $360^{\circ}$ in the figure). However, the mASRs of these two methods decreased when the viewing angle increased, which manifests the segment-missing problem. The tiled variants of both methods had some adversarial effectiveness in multiple viewing angles, while the mASRs were lower than $0.5$ in almost every viewing angles. TC-EGA outperformed the other methods at almost every viewing angle. The mASR is approximately $1.0$ at viewing angle $0^{\circ}$ and $180^{\circ}$, indicating that the person can always evade the detector when confidence threshold is larger than $0.1$. It was less effective when the viewing angle is close to $90^{\circ}$ or $270^{\circ}$ because the area captured by the camera were small at such viewing angles.

We investigated influence of the type of clothes and the distance between person and camera. From \cref{tab:threecloth}, the adversarial effectiveness varied when the texture was applied to different kinds of clothes. The attack was more effective when applying to larger clothes (e.g., dress), for more area of the texture was captured by the camera. Moreover, the adversarial clothes had comparable mASRs in both indoor and outdoor scenes (See \emph{Supplementary Materials}). Their effectiveness dropped when far from the camera (See \emph{Supplementary Materials}).

\begin{table}
\centering
\small
\begin{tabular}{lcccc}
\toprule
Clothing & Random & Tshirt & Skirt & Dress \\
\midrule
mASR       & 0.092  & 0.771  & 0.287 & 0.893   \\
\bottomrule
\end{tabular}
\captionsetup{width=\linewidth}
\captionof{table}{The mASRs of different adversarial clothes.}
\label{tab:threecloth}
\end{table}

\cref{tab:phy_masr} presents the mASRs of the adversarial clothes to attack various detectors. From the table, TC-EGA obtained much higher mASR than Random. Moreover, the adversarial effectiveness remained when the adversarial clothes are transferred across different detectors. See \emph{Supplementary Materials} for the details of the transfer study. In addition, we provide a video demo in \emph{Supplementary Video}.

\begin{table}[t]
\centering
\small
\begin{threeparttable}
\begin{tabular}{ccccc}
\toprule
Detector & YOLOv2 & YOLOv3 & FasterRCNN & MaskRCNN \\
\midrule
Random             & 0.087  & 0.000\tnote{1}  & 0.000      & 0.000   \\
TC-EGA              & 0.743  & 0.701\tnote{1}  & 0.930      & 0.855   \\
\bottomrule
\end{tabular}
\begin{tablenotes}
   \footnotesize              
   \item[1] We scaled the size of the inputs by $50\%$ before sending them to YOLOv3. See \emph{Supplementary Materials} for the reason.
\end{tablenotes}   
\captionsetup{width=\linewidth}
\captionof{table}{The mASR of different detectors in the physical world.}
\label{tab:phy_masr}
\end{threeparttable}
\end{table}

\section{Conclusions}
We propose a method to craft AdvTextures to realize physical adversarial attacks on person detection systems. The main idea is to first train a expandable generator to generate AdvTexture by taking random input in a latent space, and then search the best local patterns of the latent variable for attack. The effectiveness of the AdvTexture is improved by optimizing the latent input. We physically implemented AdvTexture by printing it on a large cloth and making different T-shirts, skirts, and dresses. Those clothes, evaluated by our experiment in the physical world, were effective when the person wearing them turns around or changes postures.

\paragraph{Limitations} Though the crafted texture targeting one detector can also attack another detector to some extent, the transferability is not very good. Model ensemble could be used to improve transferability.


\paragraph{Potential negative impact} Adversarial research may cause potentially unwanted applications in the real-world community, such as camera security issues. Many defense methods based on previously exposed vulnerabilities have been proposed~\cite{xu2017feature,goodfellow2014explaining,liao2018defense}, which have improved the security level of our community and beneficially illustrated the value of research about attack.

\label{sec:conclusion}

\section*{Acknowledgement}
This work was supported in part by the National Natural Science Foundation of China (Nos. U19B2034, 62061136001, 61836014) and the Tsinghua-Toyota Joint Research Fund.

{\small
\bibliographystyle{ieee_fullname}
\bibliography{myegbib}
}


\renewcommand\thefigure{S\arabic{figure}}
\renewcommand\thetable{S\arabic{table}}
\setcounter{figure}{0}
\setcounter{table}{0}
\onecolumn


\appendix



\section{Proofs}

\subsection{Proof of Theorem 1}

The KL divergence $\mathrm{KL}(q_\varphi(\Tilde{\tau})||p_{adv}(\Tilde{\tau}))$ can be divided into two terms:
\begin{align}
\label{eq:sup_KL}
\mathrm{KL}(q_\varphi(\Tilde{\tau})||p_{adv}(\Tilde{\tau}))=&\int_{\Tilde{\tau}} q_\varphi(\Tilde{\tau})\mathrm{log}\frac{q_\varphi(\Tilde{\tau})}{p_{adv}(\Tilde{\tau})} \dif\Tilde{\tau}\notag \\
=&\int_{\Tilde{\tau}} q_\varphi(\Tilde{\tau})\mathrm{log}{q_\varphi(\Tilde{\tau})} \dif\Tilde{\tau} - \int_{\Tilde{\tau}} q_\varphi(\Tilde{\tau})\mathrm{log}{p_{adv}(\Tilde{\tau})} \dif\Tilde{\tau},
\end{align}
where the first term is the negative entropy of $q_\varphi$, i.e., $-\mathrm{H}_\varphi(\Tilde{\tau})$. We introduce mutual information (MI) to help compute the entropy:
\begin{equation}
   \mathrm{I}_\varphi(\Tilde{\tau},z) = \int_{\Tilde{\tau}, z}{p(\Tilde{\tau},z)\mathrm{log}\frac{p(\Tilde{\tau},z)}{q_\varphi(\Tilde{\tau})p_z(z)}}\dif \Tilde{\tau} \dif z,
\end{equation}
where $p(\Tilde{\tau},z)$ is the joint distribution of $\Tilde{\tau}=G_\varphi(z)$ and $z$. Since $p(\Tilde{\tau},z) = p(\Tilde{\tau}|z)p_z(z)$ and $q_\varphi(\Tilde{\tau})$ is the marginal distribution $q_\varphi(\Tilde{\tau})=\int_z{p(\Tilde{\tau},z)\dif z}$, we have
\begin{align}
  \mathrm{I}_\varphi(\Tilde{\tau},z) 
  =&\int_{\Tilde{\tau}, z}{p(\Tilde{\tau},z)\mathrm{log}\frac{p(\Tilde{\tau},z)}{p_z(z)}}\dif \Tilde{\tau} \dif z - \int_{\Tilde{\tau}, z}{p(\Tilde{\tau},z)\mathrm{log}{q_\varphi(\Tilde{\tau})}}\dif \Tilde{\tau} \dif z\notag\\
  =&\int_{\Tilde{\tau}, z}{p(\Tilde{\tau}|z)p_z(z)\mathrm{log}p(\Tilde{\tau}|z)}\dif \Tilde{\tau} \dif z - \int_{\Tilde{\tau}}{\mathrm{log}{q_\varphi(\Tilde{\tau})}}\dif \Tilde{\tau}\int_{z}p(\Tilde{\tau},z) \dif z\notag\\
  =&\int_z p_z(z)\int_{\Tilde{\tau}}{p(\Tilde{\tau}|z)\mathrm{log}p(\Tilde{\tau}|z)}\dif \Tilde{\tau}\dif z - \int_{\Tilde{\tau}}{q_\varphi(\Tilde{\tau})\mathrm{log}{q_\varphi(\Tilde{\tau})}}\dif \Tilde{\tau}\notag\\
  =& - \mathrm{H}_\varphi(\Tilde{\tau}|z) + \mathrm{H}_\varphi(\Tilde{\tau}),
  \label{eq:info_part}
\end{align}
where $\mathrm{H}_\varphi(\Tilde{\tau}|z)$ is called conditional entropy. Therefore, the first term of \cref{eq:sup_KL} can be replaced by $-\mathrm{I}_\varphi(\Tilde{\tau},z) - \mathrm{H}_\varphi(\Tilde{\tau}|z)$. Since $\Tilde{\tau}\sim q_\varphi$ is determined by $z$, i.e., $p(\Tilde{\tau}|z) = \delta(\Tilde{\tau} - G_\varphi(z))$, we have
\begin{align}
  \mathrm{H}_\varphi(\Tilde{\tau}|z) 
  =-& \int_z p_z(z)\int_{\Tilde{\tau}}{p(\Tilde{\tau}|z)\mathrm{log}p(\Tilde{\tau}|z)}\dif \Tilde{\tau}\dif z \notag \\
  =-& \int_z p_z(z)\int_{\Tilde{\tau}}{\delta(\Tilde{\tau} - G_\varphi(z))\mathrm{log}\delta(\Tilde{\tau} - G_\varphi(z))}\dif \Tilde{\tau}\dif z \notag \\
  =-& \int_z p_z(z)\dif z\int_{\Tilde{\tau}'}{\delta(\Tilde{\tau}')\mathrm{log}\delta(\Tilde{\tau}')}\dif \Tilde{\tau}' \\
  =-& \int_{\Tilde{\tau}'}{\delta(\Tilde{\tau}')\mathrm{log}\delta(\Tilde{\tau}')}\dif \Tilde{\tau}',
\end{align}
which indicates that $\mathrm{H}_\varphi(\Tilde{\tau}|z)$ is a constant\footnote{In fact, it is zero for discrete distribution and is infinity for continuous distribution.}. Therefore, we ignore this term in \cref{eq:info_part}. Moreover, for the second term of \cref{eq:sup_KL}, since $ p_{adv}(\Tilde{\tau}) = \frac{e^{-U(\Tilde{\tau})}}{Z_U}$, we have
\begin{align}
- \int_{\Tilde{\tau}} q_\varphi(\Tilde{\tau})\mathrm{log}{p_{adv}(\Tilde{\tau})} \dif\Tilde{\tau}
=&- \int_{\Tilde{\tau}} q_\varphi(\Tilde{\tau})\mathrm{log}{\frac{e^{-U(\Tilde{\tau})}}{Z_U}} \dif\Tilde{\tau} \notag\\
=&\int_{\Tilde{\tau}} q_\varphi(\Tilde{\tau}){U(\Tilde{\tau}) \dif\Tilde{\tau} + \int_{\Tilde{\tau}} q_\varphi(\Tilde{\tau})\mathrm{log}{Z_U}} \dif\Tilde{\tau} \notag\\
=&\mathbb{E}_{\Tilde{\tau}\sim q_\varphi(\Tilde{\tau})}[U(\Tilde{\tau})] + \mathrm{log}Z_U,
\end{align}
where the partition function $Z_U=\int_{\Tilde{\tau}}{e^{-U(\Tilde{\tau})}}\mathrm{d}\Tilde{\tau}$ is a constant.

Therefore, minimizing \cref{eq:sup_KL} is equivalent to
\begin{align}
\label{ep:sup_middle}
&\min_{\varphi} -\mathrm{I}_\varphi(\Tilde{\tau},z) + \mathbb{E}_{\Tilde{\tau}\sim q_\varphi(\Tilde{\tau})}[U(\Tilde{\tau})].
\end{align}
In other words, we need to simultaneously maximize $\mathrm{I}_\varphi(\Tilde{\tau},z)$ and minimize $\mathbb{E}_{\Tilde{\tau}\sim q_\varphi(\Tilde{\tau})}[U(\Tilde{\tau})]$. According to Deep InfoMax (DIM)~\cite{hjelm2018learning}, maximizing $\mathrm{I}_\varphi(\Tilde{\tau},z)$ is equivalent to maximizing a Jensen-Shannon mutual information (MI) estimator,
\begin{align}
\mathcal{I}_{\varphi,\omega}^{\mathrm{JSD}}(\Tilde{\tau},z)=\ & \mathbb{E}_{(\Tilde{\tau}, z)\sim q_{\varphi}^{\Tilde{\tau}, z}(\Tilde{\tau}, z)}[-\mathrm{sp}(-T_{\omega}(\Tilde{\tau},z))] - \mathbb{E}_{\Tilde{\tau}\sim q_{\varphi}(\Tilde{\tau}), z'\sim p_z(z')}[\mathrm{sp}(T_{\omega}(\Tilde{\tau},z'))],
\end{align}
where $q_{\varphi}^{\Tilde{\tau}, z}$ denotes the joint distribution of $\Tilde{\tau}$ and $z$, and $\mathrm{sp}(t)=\mathrm{log}(1+e^t)$ is the softplus function. $T_\omega$ is a scalar function modeled by a neural network whose parameter $\omega$ must be optimized together with the parameter $\varphi$. Therefore, we replace $\mathrm{I}_\varphi(\Tilde{\tau},z)$ by $\mathcal{I}_{\varphi,\omega}^{\mathrm{JSD}}(\Tilde{\tau},z)$ and optimize $\varphi$ and $\omega$ simultaneously.

Given the above, minimizing $\mathrm{KL}(q_\varphi(\Tilde{\tau})||p_{adv}(\Tilde{\tau}))$ is equivalent to
\begin{equation}
\min_{\varphi, \omega} -I_{\varphi,\omega}^{\mathrm{JSD}}(\Tilde{\tau},z) + \mathbb{E}_{\Tilde{\tau}\sim q_\varphi(\Tilde{\tau})}[U(\Tilde{\tau})].
\end{equation}

\subsection{Proof of Theorem 2}

Since $G_1$ is equivalent to $G_2$, $\tau_1$ has the same dimension as $\tau_2$.We denote the dimension by $K$. Let $\tau_1^k$ be the $k$-th element of $\tau_1$, and $\tau_2^k$ be the $k$-th element of $\tau_2$. Since $\mathcal{Z}_1$ is identical to $\mathcal{Z}_2$, i.e. the probability density function (PDF) $p_{\mathcal{Z}_1}(z)=p_{\mathcal{Z}_2}(z)$, we have
\begin{align}
  \label{eq:theo2}
  &\mathrm{Pr}(\tau_1^k<h_k,k=1,2,...,K) \notag\\
  =&\int_{G_1(z)_k<h_k,k=1,2,...,K}{p_{\mathcal{Z}_1}(z)\mathrm{d}z} \notag\\
  =&\int_{G_2(z)_k<h_k,k=1,2,...,K}{p_{\mathcal{Z}_2}(z)\mathrm{d}z} \notag\\
  =&\mathrm{Pr}(\tau_2^k<h_k,k=1,2,...,K),
\end{align}
where $\{h_k\}_{k=1,2,...,K}$ is a list of arbitrary real numbers. Therefore, the cumulative distribution function (CDF) of $\mathcal{T}_1$ is equal to the CDF of $\mathcal{T}_2$, which proves that $\mathcal{T}_1$ is identical to $\mathcal{T}_2$.

\subsection{Proof of Corollary 2.1}
Assuming that the FCN has $L$ layers, we define $\mathrm{Conv}^{(l)}$, $\mathrm{Kernel}^{(l)}$ and $\mathrm{Act}^{(l)}$ as the convolutional function, the convolutional kernel and the element-wise activation function at the $l$th layer, respectively. Let the spatial size of $\mathrm{Kernel}^{(l)}$ be $a^{(l)}$ and $b^{(l)}$. We denote the value before the activation function at the $l$th layer by $o^{(l)}$ and denote the feature map by $v^{(l)}$. We further define $v^{(0)}$ as the input $z$ and define $v^{(L)}$ as the output $\tau$. Therefore, for $l\in\{1,2,...,L\}$, we have
\begin{align}
  \label{eq:conv}
  o^{(l)} &= \mathrm{Conv}^{(l)}(v^{(l-1)}) = v^{(l-1)} * \mathrm{Kernel}^{(l)}, \\
  \label{eq:act}
  v^{(l)} &= \mathrm{Act}^{(l)}(o^{(l)}),
\end{align}
where the operation $*$ stands for convolution. We denote $v^{(l)}_{i,j,w,h}$ and $o^{(l)}_{i,j,w,h}$ as a rectangular area with size $w\times h$ whose center is at the location $i,j$ in $v^{(l)}$ and $o^{(l)}$ respectively. Ignoring the boundary conditions, for all $l,i,j,i',j',w,h$, by the nature of the convolutional operation, we have
\begin{align}
  \label{eq:w1}
  o^{(l)}_{i,j,w^{(l)},h^{(l)}} &= v^{(l-1)}_{i,j,w^{(l-1)},h^{(l-1)}} * \mathrm{Kernel}^{(l)}, \\
  \label{eq:act1}
  v^{(l)}_{i,j,w^{(l)},h^{(l)}} &=  \mathrm{Act}^{(l)}(o^{(l)}_{i,j,w^{(l)},h^{(l)}}),
\end{align}
and
\begin{align}
  \label{eq:w2}
  o^{(l)}_{i',j',w^{(l)},h^{(l)}} &= v^{(l-1)}_{i',j',w^{(l-1)},h^{(l-1)}} * \mathrm{Kernel}^{(l)},\\
  \label{eq:act2}
  v^{(l)}_{i',j',w^{(l)},h^{(l)}} &=  \mathrm{Act}^{(l)}(o^{(l)}_{i',j',w^{(l)},h^{(l)}}),
\end{align}
where $w^{(l-1)} = w^{(l)} + a^{(l)} - 1$, $h^{(l-1)} = h^{(l)} + b^{(l)} - 1$, $w^{(L)}=w$, and $h^{(L)}=h$. Therefore, we can define a function $G_{i,j,w,h}^{(l)}$ as $G_{i,j,w,h}^{(l)}(v^{(0)}_{i,j,w^{(0)},h^{(0)}})=v^{(l)}_{i,j,w^{(l)},h^{(l)}}$. 

When $l = 0$, $G_{i,j,w,h}^{(l)}$ is obviously equivalent to $G_{i,j,w,h}^{(l)}$, since they are both identical functions. Moreover, the distribution of $v^{(l)}_{i,j,w^{(l)},h^{(l)}}$ is also identical to $v^{(l)}_{i',j',w^{(l)},h^{(l)}}$, since each element of $v^{(0)}$ is independent and identically distributed. 

For $l>0$, we assume that $G_{i,j,w,h}^{(l-1)}$ is equivalent to $G_{i,j,w,h}^{(l-1)}$, and the distribution of $v^{(l-1)}_{i,j,w^{(l-1)},h^{(l-1)}}$ is identical to $v^{(l-1)}_{i',j',w^{(l-1)},h^{(l-1)}}$ for all $i,j,i',j',w,h$. According to \cref{eq:w1,eq:w2,eq:act1,eq:act2}, for all $v^{(0)}_{i,j,w^{(0)},h^{(0)}}=v^{(0)}_{i',j',w^{(0)},h^{(0)}}$,
\begin{align}
   G_{i,j,w,h}^{(l)}(v^{(0)}_{i,j,w^{(0)},h^{(0)}}) &= \mathrm{Act}^{(l)}(v^{(l-1)}_{i,j,w^{(l-1)},h^{(l-1)}} * \mathrm{Kernel}^{(l)})\notag \\
   &= \mathrm{Act}^{(l)}(G_{i,j,w,h}^{(l-1)}(v^{(0)}_{i,j,w^{(0)},h^{(0)}})) * \mathrm{Kernel}^{(l)})\notag \\
   &= \mathrm{Act}^{(l)}(G_{i',j',w,h}^{(l-1)}(v^{(0)}_{i',j',w^{(0)},h^{(0)}})) * \mathrm{Kernel}^{(l)})\notag \\
   &= \mathrm{Act}^{(l)}(v^{(l-1)}_{i',j',w^{(l-1)},h^{(l-1)}} * \mathrm{Kernel}^{(l)})\notag \\
   &= G_{i',j',w,h}^{(l)}(v^{(0)}_{i',j',w^{(0)},h^{(0)}}).\notag
\end{align}
Therefore, $G_{i,j,w,h}^{(l)}$ is equivalent to $G_{i',j',w,h}^{(l)}$. Moreover, since the convolutions in \cref{eq:w1,eq:w2} are equivalent, the distribution of $o^{(l)}_{i,j,w^{(l)},h^{(l)}}$ is identical to that of $o^{(l)}_{i',j',w^{(l)},h^{(l)}}$ according to Theorem 2. Furthermore, Since the active function $\mathrm{Act}^{(l)}$ is element-wise, i.e., it is equivalent for $i,j$ and $i',j'$, the distribution of $v^{(l)}_{i,j,w^{(l)},h^{(l)}}$ is also identical to that of $v^{(l)}_{i',j',w^{(l)},h^{(l)}}$ according to Theorem 2.

By using mathematical induction, we conclude that $G_{i,j,w,h}^{(l)}$ is equivalent to $G_{i',j',w,h}^{(l)}$, and the distribution of $v^{(l)}_{i,j,w^{(l)},h^{(l)}}$ is identical to that of $v^{(l)}_{i',j',w^{(l)},h^{(l)}}$ for all $l\in[1,2,...,L]$. Since $v^{(L)}=\tau$, $w^{(l)}=w$, and $h^{(l)}=h$, we derive Corollary 2.1. Note that every convolutional layer of FCN needs to be zero-padded to avoid the boundary problem.

\section{Adversarial Textures and Adversarial Clothes}

\begin{figure}[h]
    \centering
    \begin{subfigure}{0.15\linewidth}
    \includegraphics[width=\textwidth]{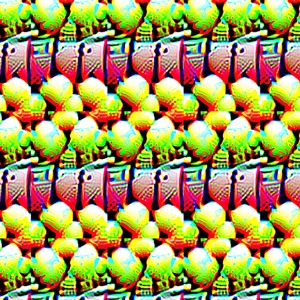}
    \caption{AdvTshirtTile}
    \label{fig:AdvTshirtTile}
    \end{subfigure}
    \begin{subfigure}{0.15\linewidth}
    \includegraphics[width=\textwidth]{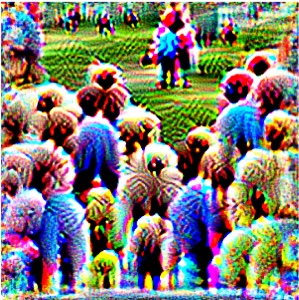}
    \caption{RCA2$\times$}
    \label{fig:rca2}
    \end{subfigure}
    \begin{subfigure}{0.15\linewidth}
    \includegraphics[width=\textwidth]{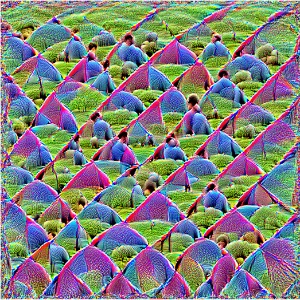}
    \caption{RCA6$\times$}
    \label{fig:rca6}
    \end{subfigure}
    \begin{subfigure}{0.15\linewidth}
    \includegraphics[width=\textwidth]{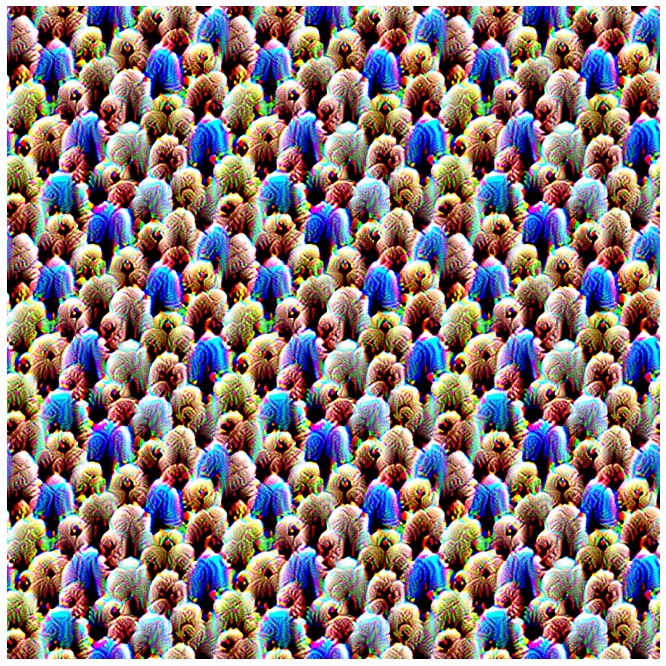}
    \caption{TCA}
    \label{fig:tca}
    \end{subfigure}

    \begin{subfigure}{0.15\linewidth}
    \includegraphics[width=\textwidth]{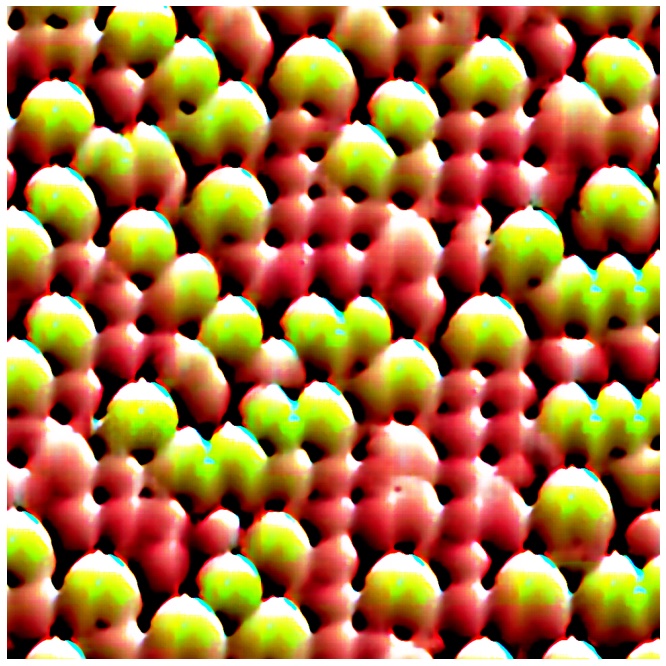}
    \caption{EGA}
    \label{fig:ega}
    \end{subfigure}
    \begin{subfigure}{0.15\textwidth}
    \includegraphics[width=\textwidth]{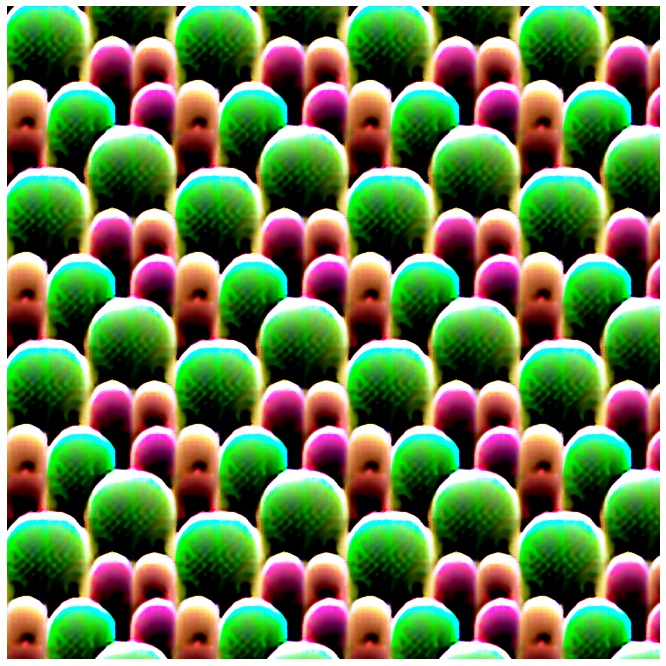}
    \caption{YOLOv3~\cite{redmon2018yolov3}}
    \label{fig:v3}
    \end{subfigure}
    \begin{subfigure}{0.15\textwidth}
    \includegraphics[width=\textwidth]{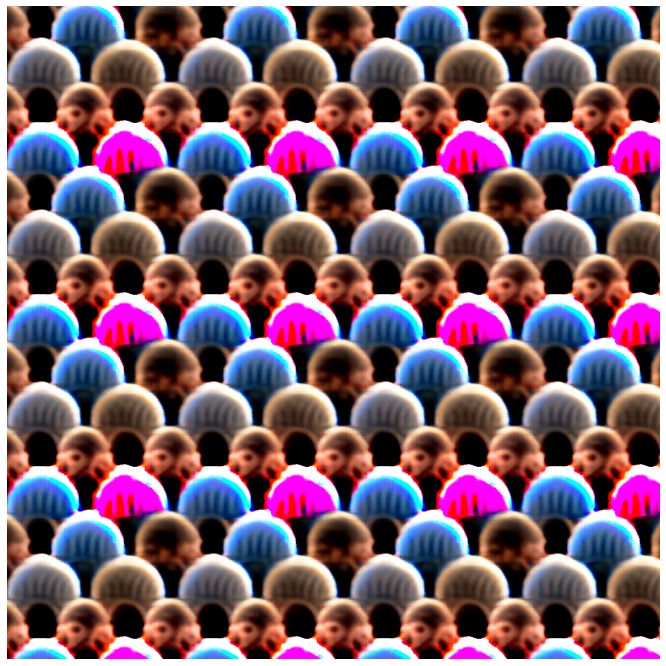}
    \caption{FasterRCNN~\cite{ren2016faster}}
    \label{fig:fr}
    \end{subfigure}
    \begin{subfigure}{0.15\linewidth}
    \includegraphics[width=\textwidth]{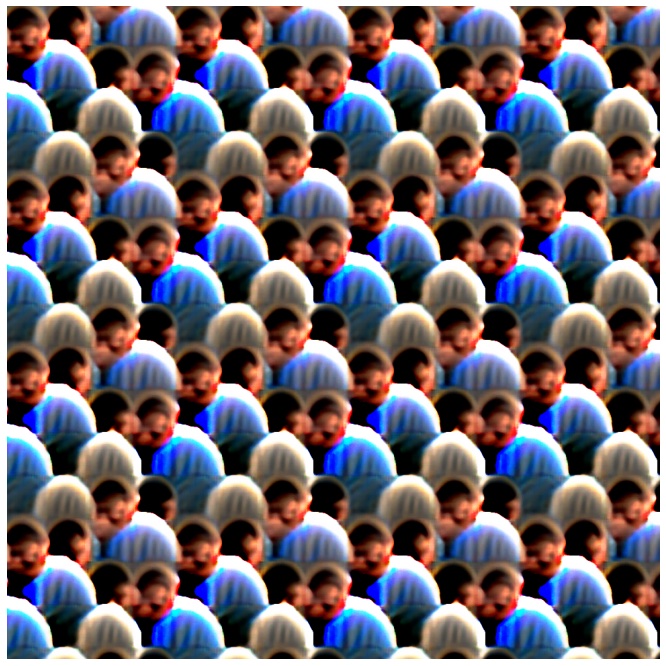}
    \caption{MaskRCNN~\cite{he2017mask}}
    \label{fig:mr}
    \end{subfigure}

    \caption{Visualization of different adversarial textures, extending Fig.~6 in the main paper. (a) The texture formed by tiling an adversarial patches~\cite{xu2020adversarial} repeatedly. (b-e) The textures produced by different methods to attack YOLOv2~\cite{redmon2017yolo9000}. (f-h) The textures produced by TC-EGA to attack different detectors respectively.}
    \label{fig:pattern_vis_sup}
\end{figure}

\begin{figure}[h]
\centering
\includegraphics[width=0.6\textwidth]{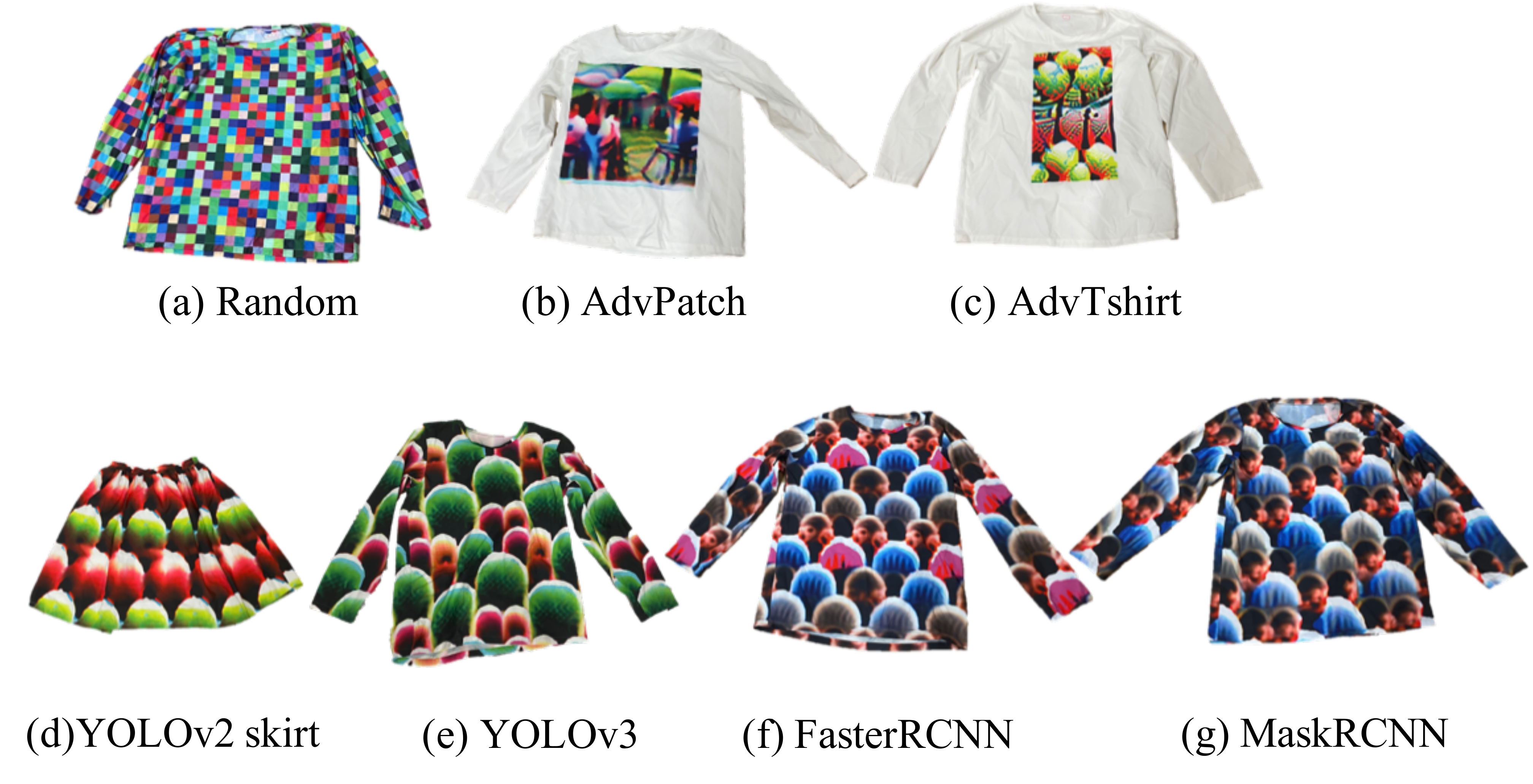}
\caption{Real-world adversarial clothes produced by different methods, extending Fig.~8 in the main paper.}
\label{fig:clothes_sup}
\end{figure}

\cref{fig:pattern_vis_sup,fig:clothes_sup} present additional adversarial textures and adversarial clothes, respectively, that are not presented in the main paper (Figs.~6 and 8) due to the page limit. Unless otherwise specified, all results about physical attacks presented in both the main paper and the \emph{Supplementary Materials} were obtained by adversarial T-shirts. 

\section{Results of attacking different detectors in the digital world}

\begin{table}
\centering
\small
\begin{tabular}{cccc}
\toprule
Target detector & YOLOv3 & FasterRCNN & MaskRCNN \\
\midrule
AP              & 0.511  & 0.419      & 0.492   \\
\bottomrule
\end{tabular}
\captionsetup{width=\linewidth}
\captionof{table}{The APs of different detectors attacked by TC-EGA on Inria test set.}
\label{tab:patch_other}
\end{table}

\cref{tab:patch_other} presents the APs of YOLOv3, FasterRCNN and MaskRCNN on Inria test set. Note that the AP of each detector on the original test images is $1.0$. Though these AdvTextures were not effective as that of YOLOv2 whose AP was $0.362$ (See Tab. 1 in the main paper), they had lowered the AP of clean images by half.

\section{Comparison between Indoor and Outdoor Conditions}

We compared the attack effectiveness of different adversarial T-shirts in the indoor and outdoor scenes. We used the videos described in Sec.4.2 in the main paper. We extracted $32$ frames from each video with viewing angles varying from $0^{\circ}$ to $3^{\circ}$. Therefore we collected $3\times32=96$ frames for each scene and each detector. The results are presented in \cref{tabfig:white_split}. The indoor mASR was comparable to the outdoor mASR for each piece of adversarial clothing. It indicates that the adversarial clothes are effective in different scenes.


\begin{table}[h]
\centering
\small
\begin{tabular}{lcccc}
\hline
\diagbox{scene}{Target} & YOLOv2 & YOLOv3 & FasterRCNN & MaskRCNN  \\
\hline
Indoor          & 0.771    & 0.764    & 0.912    & 0.832        \\
Outdoor          & 0.714    & 0.638    & 0.948    & 0.878       \\
\bottomrule
\end{tabular}
\caption{The mASRs of the attacks at different distances between persons and camera.}
\label{tabfig:white_split}
\end{table}


\section{Effectiveness of the Attack with Respect to the Distance to the Camera}

\begin{figure}[h]
    \centering
    \includegraphics[width=.3\textwidth]{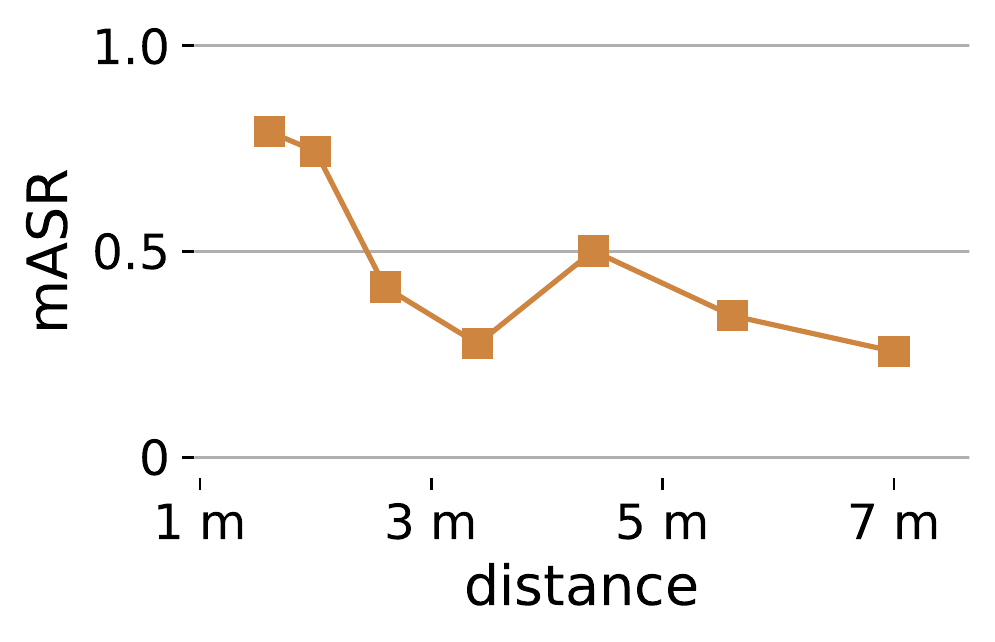}
    \caption{The mASRs of the attacks at different distances between persons and camera.}
    \label{fig:dis}
\end{figure}

We recorded additional videos for each person wearing YOLOv2 T-shirt in both indoor and outdoor scenes. The persons still turned a circle slowly in front of the camera to collect frames at different viewing angles. We varied the distance between the camera and the persons to be $1.6\;\mathrm{m}$, $2.0\;\mathrm{m}$, $2.6\;\mathrm{m}$, $3.4\;\mathrm{m}$, $4.4\;\mathrm{m}$, $5.6\;\mathrm{m}$, and $7.0\;\mathrm{m}$. For each distance, we collected $3\text{(persons)}\times2\text{(scenes)}\times32\text{(frames per video)}=192$ frames in total. \cref{fig:dis} presents the mASRs of YOLOv2 T-shirt at various distances. The mASR was the highest when the persons was close to the camera ($1.6\;\mathrm{m}$, mASR $0.791$). It decreased to $0.257$ when the distance was $7.0\;\mathrm{m}$. 

\section{Attacking YOLOv3}

\begin{figure}[h]
\centering
\begin{subfigure}{.32\linewidth}
  \includegraphics[height=1.3in]{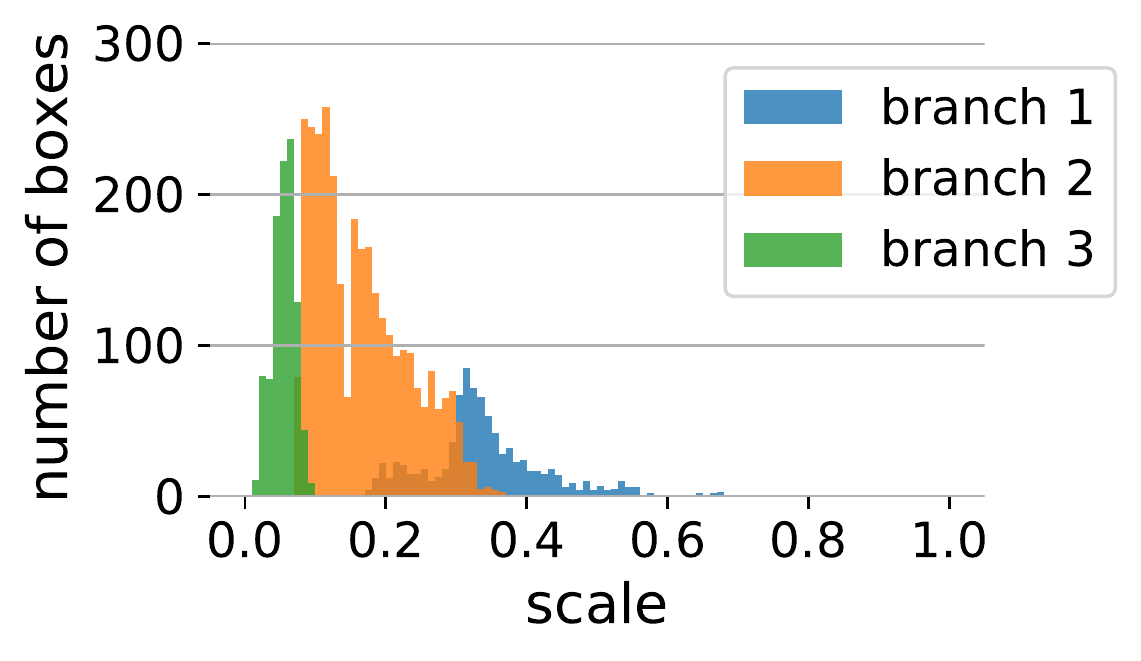}
  \caption{}
  \label{fig:hist_split}
\end{subfigure}
\begin{subfigure}{.38\linewidth}
  \includegraphics[height=1.2in]{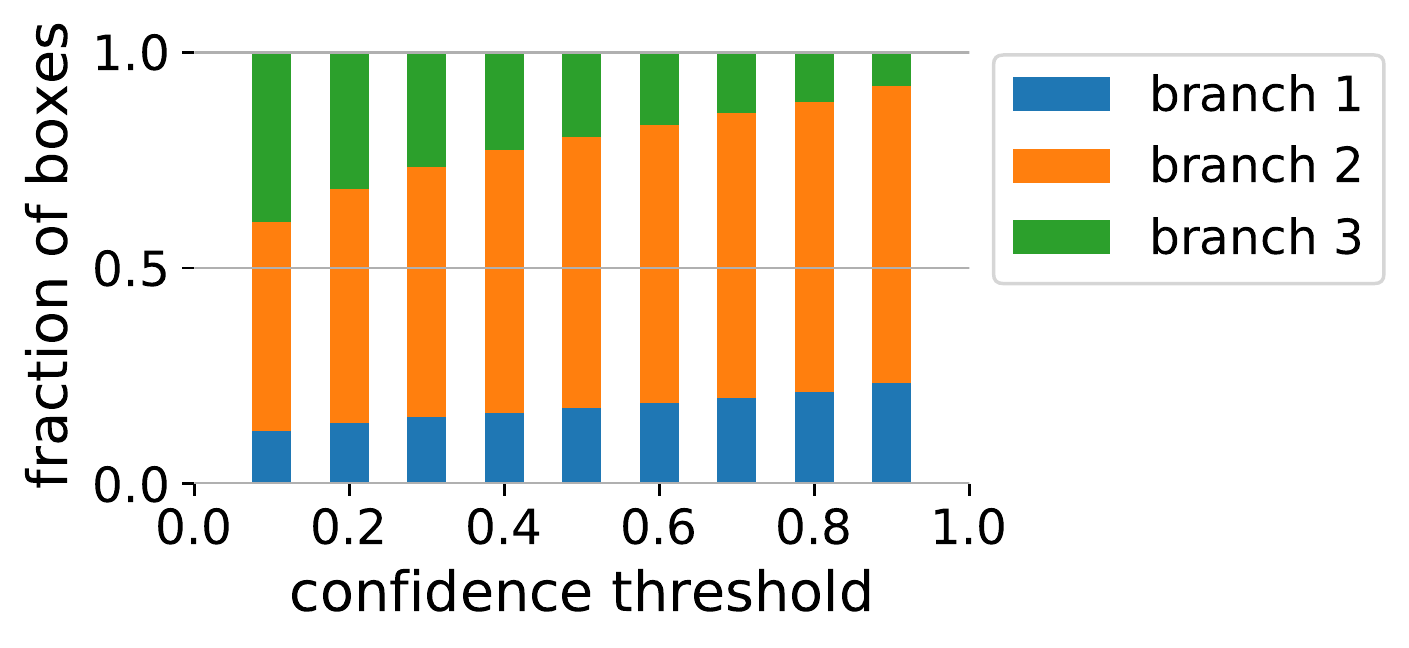}
  \caption{}
  \label{fig:fraction_split}
\end{subfigure}
\begin{subfigure}{.28\linewidth}
  \includegraphics[height=1.3in]{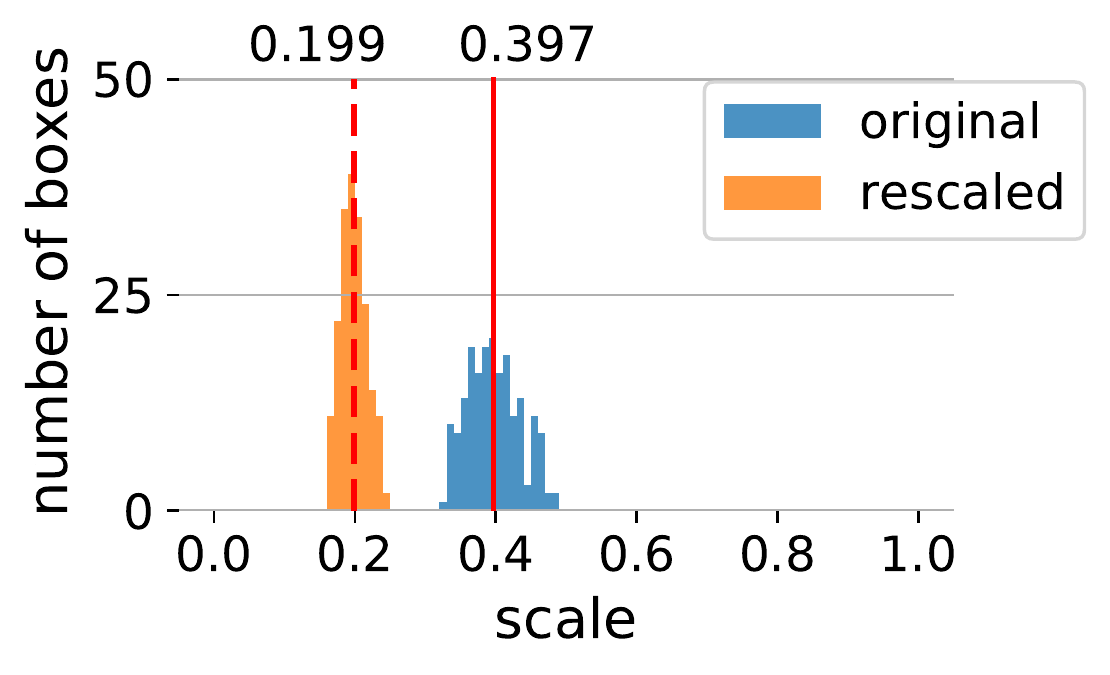}
  \caption{}
  \label{fig:video_hist}
\end{subfigure}
\caption{(a) The distribution of the boxes' scales predicted by different branches of YOLOv3. For each box with normalized size $w\times h$, we define the scale by $\sqrt{w*h}$. (b) The fractions of the boxes predicted by different branches with respect to various confidence thresholds. (c) The distribution of the scales of the boxes on the original and rescaled video frames. The red solid line denotes the average scale on the original video frames, and the red dashed line denotes the average scale on the rescaled frames (by $50\%$).}
\label{fig:split_v3}
\end{figure}

In this section we provide the reasons of scaling the size of the input by $50\%$ before sending to YOLOv3 (see Tab.~4 in the main paper). YOLOv3 has three branches to predict boxes in different scales. These branches are based on feature maps of an backbone network in different layers, and use additional blocks before predicting boxes. Therefore, These branches are relatively independent when being adversarially attacked. Since the number of the boxes predicted by different branches can be quite different, the attack might be biased to one particular branch. \cref{fig:hist_split} presents the histogram of the predicted boxes of each branch on the Inria training dataset, with a confidence threshold $0.5$. The first branch predicted large scale boxes, and the third predicted small scale boxes. \cref{fig:fraction_split} presents the fraction of the predicted boxes with respect to different confidence thresholds. From the figure, the second branch predicted most of the boxes ($62.8\%$ when the confidence threshold is $0.5$), indicating that the produced adversarial pattern may be biased towards attacking the second branch. However, in our recorded videos, the scale of the persons were outside the range of the second branch's predicted boxes (compare \cref{fig:hist_split,fig:video_hist}). Therefore, we scaled the size of the input by $50\%$ before sending the frames to YOLOv3.

\section{Transfer Study in the Physical World}
We performed transfer-based attacks on several detectors by the adversarial clothes that are produced to attack particular detectors. \cref{tab:transfer} presents the mASR of the transfer-based attacks. Every number in the table was obtained over $192$ frames as described in Section 4.2 in the main paper. The adversarial clothes of YOLOv2 and YOLOv3 remained effective when they were used to attack YOLOv3 and YOLOv2, respectively. However, these clothes got low mASRs when attacking other models except RetinaNet. The adversarial clothes of Faster RCNN and MaskRCNN remained effective when they were used to attack other models, though sometimes (e.g., attacking YOLOv3) not as effective as attacking themselves. A possible solution is to use the model ensemble technique \cite{dong2018boosting,liu2017delving}, which is left as future research.

\begin{table}[h]
\centering
\small
\begin{tabular}{lccccccc}
\hline
\diagbox{source}{target} & YOLOv2 & YOLOv3 & FasterRCNN & MaskRCNN & RetinaNet~\cite{lin2017focal} & Cascade MaskRCNN~\cite{cai2019cascade} & \\
\hline
YOLOv2          & 0.743    & 0.526    & 0.000    & 0.000    & 0.182    & 0.000     \\
YOLOv3          & 0.518    & 0.701    & 0.014    & 0.037    & 0.453    & 0.009     \\
FasterRCNN      & 0.617    & 0.237    & 0.930    & 0.848    & 0.900    & 0.695     \\
MaskRCNN        & 0.547    & 0.359    & 0.873    & 0.855    & 0.838    & 0.575     \\
\bottomrule
\end{tabular}
\caption{The mASRs of transferred attack.}
\label{tab:transfer}
\end{table}

\end{document}